\pdfoutput=1
\documentclass[journal]{IEEEtran}
\usepackage[T1]{fontenc}
\usepackage[latin9]{inputenc}
\usepackage{color}
\usepackage{array}
\usepackage{float}
\usepackage{calc}
\usepackage{amsmath}
\usepackage{amsthm}
\usepackage{amssymb}
\usepackage{graphicx}
\usepackage[unicode=true,
 bookmarks=true,bookmarksnumbered=true,bookmarksopen=true,bookmarksopenlevel=1,
 breaklinks=false,pdfborder={0 0 0},pdfborderstyle={},backref=false,colorlinks=false]
 {hyperref}
\hypersetup{pdftitle={Your Title},
 pdfauthor={Your Name},
 pdfpagelayout=OneColumn, pdfnewwindow=true, pdfstartview=XYZ, plainpages=false}

\makeatletter

\providecommand{\tabularnewline}{\\}
\floatstyle{ruled}
\newfloat{algorithm}{tbp}{loa}
\providecommand{\algorithmname}{Algorithm}
\floatname{algorithm}{\protect\algorithmname}

\let\oldforeign@language\foreign@language
\DeclareRobustCommand{\foreign@language}[1]{%
  \lowercase{\oldforeign@language{#1}}}
\theoremstyle{plain}
\newtheorem{thm}{\protect\theoremname}
\theoremstyle{plain}
\newtheorem{lem}[thm]{\protect\lemmaname}

\usepackage{algorithm,algpseudocode}

\@ifundefined{showcaptionsetup}{}{%
 \PassOptionsToPackage{caption=false}{subfig}}
\usepackage{subfig}
\makeatother

\providecommand{\lemmaname}{Lemma}
\providecommand{\theoremname}{Theorem}

\begin{document}
\title{Structured Bayesian Compression for Deep Neural Networks Based on
The Turbo-VBI Approach}
\author{Chengyu~Xia, Danny H. K. Tsang,~\IEEEmembership{Fellow,~IEEE,}
and Vincent K. N.~Lau$^{*}$,~\IEEEmembership{Fellow,~IEEE}\thanks{$^{*}$Vincent Lau is the corresponding author.\protect \\
Chengyu~Xia, Danny Tsang and Vincent Lau are with the Department
of Electronic and Computer Engineering, The Hong Kong University of
Science and Technology, Hong Kong, e-mail: \protect\href{mailto:cxiaab@connect.ust.hk}{cxiaab@connect.ust.hk};
\protect\href{mailto:eetsang@ece.ust.hk}{eetsang@ece.ust.hk}; \protect\href{mailto:eeknlau@ece.ust.hk}{eeknlau@ece.ust.hk}.}}
\markboth{Journal of XXX}{Chengyu Xia \MakeLowercase{et al.}: Structured Bayesian Compression
for Deep Neural Networks Based on Turbo-VBI Approach}
\maketitle
\begin{abstract}
With the growth of neural network size, model compression has attracted
increasing interest in recent research. As one of the most common
techniques, pruning has been studied for a long time. By exploiting
the structured sparsity of the neural network, existing methods can
prune neurons instead of individual weights. However, in most existing
pruning methods, surviving neurons are randomly connected in the neural
network without any structure, and the non-zero weights within each
neuron are also randomly distributed. Such irregular sparse structure
can cause very high control overhead and irregular memory access for
the hardware and even increase the neural network computational complexity.
In this paper, we propose a three-layer hierarchical prior to promote
a more regular sparse structure during pruning. The proposed three-layer
hierarchical prior can achieve per-neuron weight-level structured
sparsity and neuron-level structured sparsity. We derive an efficient
Turbo-variational Bayesian inferencing (Turbo-VBI) algorithm to solve
the resulting model compression problem with the proposed prior. The
proposed Turbo-VBI algorithm has low complexity and can support more
general priors than existing model compression algorithms. Simulation
results show that our proposed algorithm can promote a more regular
structure in the pruned neural networks while achieving even better
performance in terms of compression rate and inferencing accuracy
compared with the baselines.
\end{abstract}


\IEEEpeerreviewmaketitle{}

\section{Introduction\label{sec:Introduction}}

\IEEEPARstart{D}{eep} neural networks (DNNs) have been extremely
successful in a wide range of applications. However, to achieve good
performance, state-of-the-art DNNs tend to have huge numbers of parameters.
Deployment and transmission of such big models pose significant challenges
for computation, memory and communication. This is particularly important
for edge devices, which have very restricted resources. For the above
reasons, model compression has become a hot topic in deep learning.

A variety of research focused on model compression exists. In \cite{Lee2018}\nocite{Lee2018,LeCun1990,Hassibi1993}-\cite{Hassibi1993},
the weight pruning approaches are adopted to reduce the complexity
of DNN inferencing. In \cite{Lee2018}, the threshold is set to be
the weight importance, which is measured by introducing an auxiliary
importance variable for each weight. With the auxiliary variables,
the pruning can be performed in one shot and before the formal training.
In \cite{LeCun1990} and \cite{Hassibi1993}, Hessian matrix of the
weights are used as a criterion to perform pruning. These aforementioned
approaches, however, cannot proactively force some of the less important
weights to have small values and hence the approaches just passively
prune small weights. In \cite{Louizos2017a}\nocite{Wen2016,Neill2020,Scardapane2017,howard2019searching,zhuang2020neuron,yang2020harmonious,Weigend1991,frankle2018lottery}-\cite{frankle2020linear},
systematic approaches of model compression using optimization of regularized
loss functions are adopted. In \cite{Weigend1991}, an $\ell_{2}$-norm
based regularizer is applied to the weights of a DNN and the weights
that are close to zero are pruned. In \cite{Louizos2017a}, an $\ell_{0}$-norm
regularizer, which forces the weights to be exactly zero, is proposed.
The $\ell_{0}$-norm regularization is proved to be more efficient
to enforce sparsity in the weights but the training problem is notoriously
difficult to solve due to the discontinuous nature of the $\ell_{0}$-norm.
In these works, the regularization only promotes sparsity in the weights.
Yet, sparsity in the weights is not equivalent to sparsity in the
neurons. In \cite{zhuang2020neuron}, polarization regularization
is proposed, which can push a proportion of weights to zero and others
to values larger than zero. Then, all the weights connected to the
same neuron are assigned with a common polarization regularizer, such
that some neurons can be entirely pruned and some can be pushed to
values larger than zero. In this way, the remaining weights do not
have to be pushed to small values and hence can improve the expressing
ability of the network. In \cite{Wen2016}, the authors use a group
Lasso regularizer to remove entire filters in CNNs and show that their
proposed group sparse regularization can even increase the accuracy
of a ResNet. However, the resulting neurons are randomly connected
in the neural network and the resulting datapath of the pruned neural
network is quite irregular, making it difficult to implement in hardware
\cite{Wen2016,Neill2020}.

In addition to the aforementioned deterministic regularization approaches,
we can also impose sparsity in the weights of the neural network using
Bayesian approaches. In \cite{van2020bayesian}, weight pruning and
quantization are realized at the same time by assigning a set of quantizing
gates to the weight value. A prior is further designed to force the
quantizing gates to ``close'', such that the weights are forced
to zero and the non-zero weights are forced to a low bit precision.
By designing an appropriate prior distribution for the weights, one
can enforce more refined structures in the weight vector. In \cite{Molchanov2017},
a simple sparse prior is proposed to promote weight-level sparsity,
and variational Bayesian inference (VBI) is adopted to solve the model
compression problem. In \cite{Louizos2017}, group sparsity is investigated
from a Bayesian perspective. A two-layer hierarchical sparse prior
is proposed where the weights follow Gaussian distributions and all
the output weights of a neuron share the same prior variance. Thus,
the output weights of a neuron can be pruned at the same time and
neuron-level sparsity is achieved.

In this paper, we consider model compression of DNNs from a Bayesian
perspective. Despite various existing works on model compression,
there are still several technical issues to be addressed.
\begin{itemize}
\item \textbf{Structured datapath in the pruned neural network:} In deterministic
regularization approaches, we can achieve group sparsity in a weight
matrix. For example, under $\ell_{2,1}$-norm regularization \cite{Wen2016,Scardapane2017},
some rows can be zeroed out in the weight matrix after pruning and
this results in neuron-level sparsity. However, surviving neurons
are randomly connected in the neural network without any structure.
Moreover, the non-zero weights within each neuron are also randomly
distributed. In the existing Bayesian approaches, both the single-layer
priors in \cite{Molchanov2017} and the two-layer hierarchical priors
in \cite{Louizos2017} also cannot promote structured neuron connectivity
in the pruned neural network. As such, the random and irregular neuron
connectivity in the datapath of the DNN poses challenges in the hardware
implementation.\footnote{When the pruned DNN has randomly connected neurons and irregular weights
within each neuron, very high control overhead and irregular memory
access will be involved for the datapath to take advantage of the
compression in the computation of the output.}
\item \textbf{Efficiency of weight pruning: }With existing model compression
approaches, the compressed model can still be too large for efficient
implementation on mobile devices. Moreover, the existing Bayesian
model compression approaches tend to have high complexity and low
robustness with respect to different data sets, as they typically
adopt Monte Carlo sampling in solving the optimizing problem \cite{Molchanov2017,Louizos2017}.
Thus, a more efficient model compression solution that can achieve
a higher compression rate and lower complexity is required.
\end{itemize}
To overcome the above challenges, we propose a three-layer hierarchical
sparse prior that can exploit structured weight-level and neuron-level
sparsity during training. To handle the hierarchical sparse prior,
we propose a Turbo-VBI \cite{Liu2020} algorithm to solve the Bayesian
optimization problem in the model compression. The main contributions
are summarized below.
\begin{itemize}
\item \textbf{Three-layer Hierarchical Prior for Structured Weight-level
and Neuron-level Sparsity: }The design of an appropriate prior distribution
in the weight matrix is critical to achieve more refined structures
in the weights and neurons. To exploit more structured weight-level
and neuron-level sparsity, we propose a three-layer hierarchical prior
which embraces both of the traditional two-layer hierarchical prior
\cite{Louizos2017} and support-based prior \cite{Schniter2010}.
The proposed three-layer hierarchical prior has the following advantages
compared with existing sparse priors in Bayesian model compression:
1) It promotes advanced weight-level structured sparsity as well as
neuron-level structured sparsity, as illustrated in Fig. \ref{fig:Proposed-structure}.
Unlike existing sparse priors, our proposed prior not only prunes
the neurons, but also promotes regular structures in the unpruned
neurons as well as the weights of a neuron. Thus, it can achieve more
regular structure in the overall neural network. 2) It is flexible
to promote different sparse structures. The group size as well as
the average gap between two groups can be adjusted via tuning the
hyper parameters. 3) It is optimizing-friendly. It allows the application
of low complexity training algorithms.
\item \textbf{Turbo-VBI Algorithm for Bayesian Model Compression:} Based
on the 3-layer hierarchical prior, the model training and compression
is formulated as a Bayesian inference problem. We propose a low complexity
Turbo-VBI algorithm with some novel extensions compared to existing
approaches: 1) It can support more general priors. Existing Bayesian
compression methods \cite{Molchanov2017,Louizos2017} cannot handle
the three-layer hierarchical prior due to the extra layer in the prior.
2) It has higher robustness and lower complexity compared with existing
Bayesian compression methods. Existing Bayesian compression methods
use Monte Carlo sampling to approximate the log-likelihood term in
the objective function, which has low robustness and high complexity
\cite{Wu2018}. To overcome this drawback, we propose a deterministic
approximation for the log-likelihood term.
\item \textbf{Superior Model Compression Performance:} The proposed solution
can achieve a highly compressed neural network compared to the state-of-the-art
baselines and achieves a similar inferencing performance. Therefore,
the proposed solution can substantially reduce the computational complexity
in the inferencing of the neural network and the resulting pruned
neural networks are hardware friendly. Consequently, it can fully
unleash the potential of AI applications over mobile devices.
\end{itemize}
The rest of the paper is organized as follows. In Section \ref{sec:desired-structures-in-weight-matrix},
the existing structures and the desired structure in the weight matrix
are elaborated. The three-layer hierarchical prior for the desired
structure is also introduced. In Section \ref{sec:problem-formulation},
the model compression problem is formulated. In Section \ref{sec:turbo-vbi-algorithm},
the Turbo-VBI algorithm for model compression under the three-layer
hierarchical prior is presented. In Section \ref{sec:Performance-Analysis}
and Section \ref{sec:Conclusions}, numerical experiments and conclusions,
respectively, are provided.

\section{Structures in The Weight Matrix\label{sec:desired-structures-in-weight-matrix}}

Model compression is realized by enforcing sparsity in the weight
matrix. We first review the main existing structured sparsity of the
weight matrix in the literature. Based on this, we elaborate the desired
structures in the weight matrix that are not yet considered. Finally,
we propose a 3-layer hierarchical prior model to enforce such a desired
structure in the weight matrix.

\subsection{Review of Existing Sparse Structured Weight Matrix}
\begin{center}
\begin{figure}[tb]
\subfloat[\label{fig:Random-sparsity-structure}Random sparsity structure.]{\begin{centering}
\includegraphics[width=0.45\textwidth]{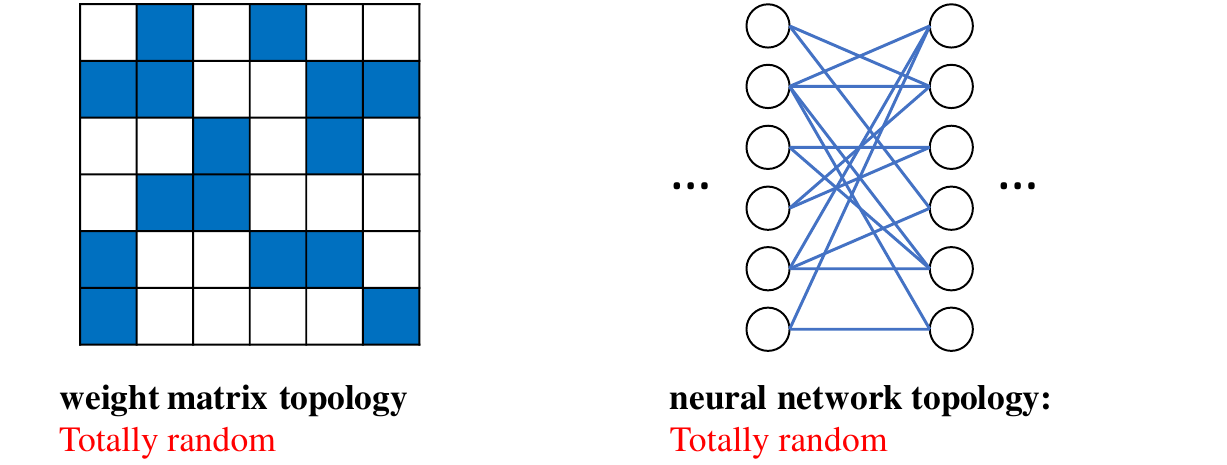}
\par\end{centering}
}\hfill{}\subfloat[Group sparsity structure.\label{fig:Group-sparsity-structure}]{\begin{centering}
\includegraphics[width=0.45\textwidth]{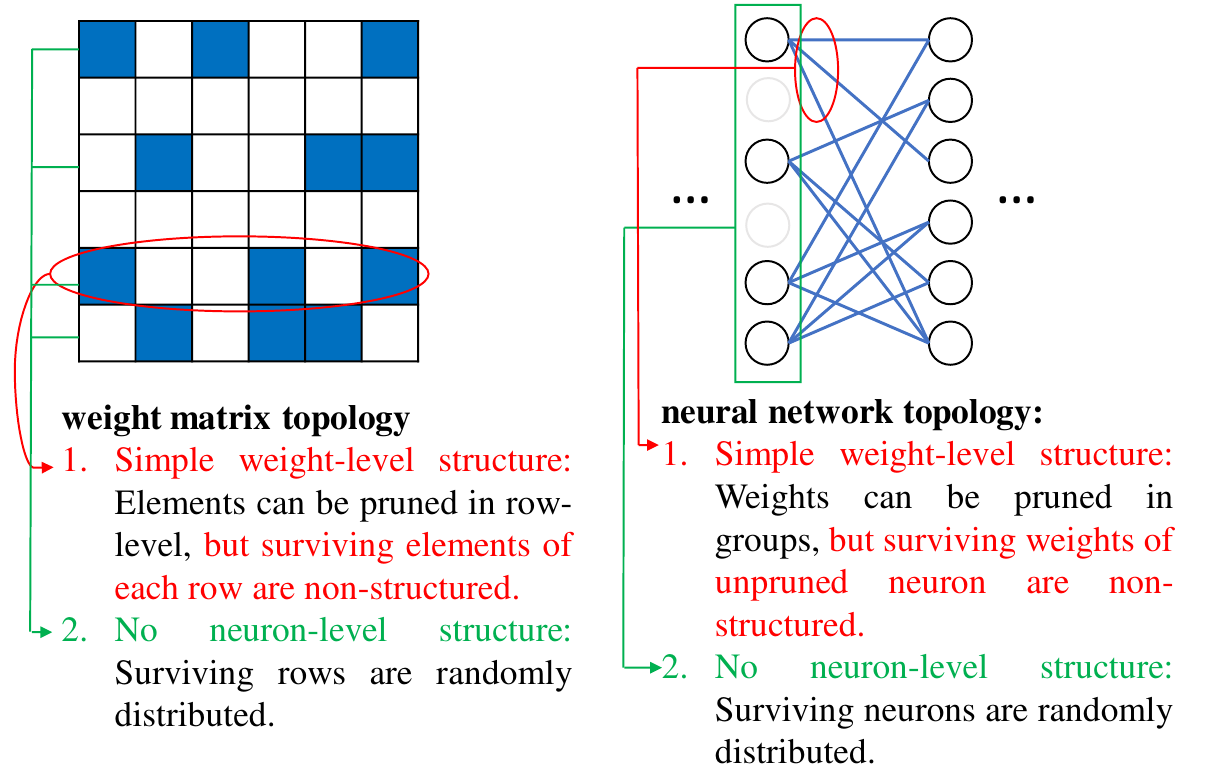}
\par\end{centering}
}\hfill{}\subfloat[Proposed multi-level structure.\label{fig:Proposed-structure}]{\centering{}\includegraphics[width=0.45\textwidth]{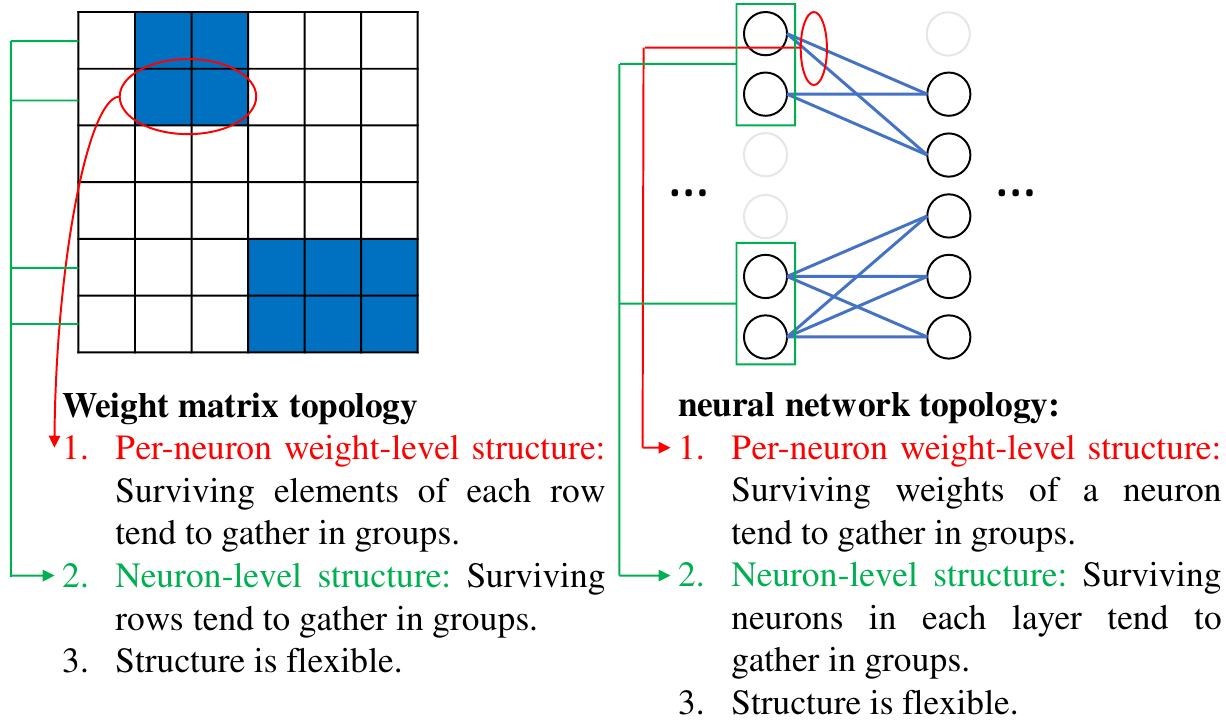}}

\caption{Comparison of different structures in the weight matrix of a fully
connected layer with 6 input neurons and 6 output neurons. Each row
denotes the weights connected to one input neuron.}
\end{figure}
\par\end{center}

There are two major existing sparse structures in the weight matrix,
namely random sparsity and group sparsity. We focus on a fully connected
layer to elaborate these two sparse structures.

\textbf{1) Random Sparsity in the Weight Matrix:} In random sparsity,
the weights are pruned individually. $\ell_{1}$-norm regularization
and one-layer prior \cite{Molchanov2017} are often adopted to regularize
individual weights. Such regularization is simple and optimization-friendly;
however, both resulting weights and neurons are completely randomly
connected since there is no extra constraint on the pruned weights.
In the weight matrix, this means that the elements are set to zero
without any structure, which results in the remaining elements also
having no specific structures. When translated into neural network
topology, this means that the surviving weights and neurons are randomly
connected, as illustrated in Fig. \ref{fig:Random-sparsity-structure}.
Because of the random connections in the pruned network, such a structure
is not friendly to practical applications such as storage, computation
and transmission \cite{Wen2016,zhang2019eager}.

\textbf{2) Group Sparsity in the Weight Matrix:} In group sparsity,
the weights are pruned in groups. Particularly, in fully connected
layers, a group is often defined as the outgoing weights of a neuron
to promote neuron-level sparsity. To promote group sparsity, sparse
group Lasso regularization \cite{Scardapane2017,Friedman2010} and
two-layer priors are proposed. Generally, such regularization often
has two folds: one regularizer for group pruning and one regularizer
for individual weight pruning. Thus, the weights can not only be pruned
in groups, but can also be pruned individually in case the group cannot
be totally removed. However, when a neuron cannot be totally pruned,
the weights connected to it are randomly pruned, and the neurons are
also pruned in random. In Fig. \ref{fig:Group-sparsity-structure},
we illustrate an example of group sparsity. In the weight matrix,
the non-zero elements in a row are randomly distributed and the non-zero
rows are also randomly distributed. When translated into neural network
topology, this means the unpruned weights connected to a neuron are
randomly distributed, and in each layer, the surviving neurons are
also randomly distributed. Thus, group sparsity is still not fully
structured.

On the other hand, recent research has pointed out that the irregular
connections in the neural network can bring various disadvantages
in reducing the computational complexity or inferencing time \cite{zhang2019eager},
\cite{yang2017designing}-\nocite{yang2017designing,kwon2020structured,Yu2017,blalock2020state}\cite{blalock2020state}.
As discussed above, existing sparsity structures are not regular enough,
and the drawbacks of such irregular structures are listed as follows.
\begin{enumerate}
\item \emph{Low decoding and computing parallelism:} Sparse weight matrix
are usually stored in a sparse format on the hardware, e.g., compressed
sparse row (CSR), compressed sparse column (CSC), coordinate format
(COO). However, as illustrated in Fig. \ref{fig:pros and cons} (a),
when decoding from the sparse format, the number of decoding steps
for each row (column) can vastly differ due to the irregular structure.
This can greatly harm the decoding parallelism \cite{kwon2020structured}.
Moreover, the irregular structure also hampers the parallel computation
such as matrix tiling \cite{Yu2017,zhang2019eager}, as illustrated
in Fig. \ref{fig:pros and cons} (b).
\item \emph{Inefficient pruning of neurons:} As illustrated in Fig. \ref{fig:pros and cons}
(c), the unstructured surviving weights from the previous layer tend
to activate random neurons in the next layer, which is not efficient
in pruning the neurons in the next layer. This may lead to inefficiency
in reducing the floating point operations (FLOPs) of the pruned neural
network.
\item \emph{Large storage burden:} Since the surviving weights of existing
model compression methods are unstructured, the exact location for
each surviving weight has to be recorded. This leads to a huge storage
overhead, which can be even greater than storing the dense matrix
\cite{Yu2017,zhang2019eager}. The huge storage overhead can also
affect the inferencing time by increasing the memory access and decoding
time.
\end{enumerate}

\subsection{Multi-level Structures in the Weight Matrix}

Based on the above discussion, a multi-level structured sparse neural
network should meet the following three criteria: \textbf{1) Per-neuron
weight-level structured sparsity:} The surviving weights connected
to a neuron should exhibit a regular structure rather than be randomly
distributed as in traditional group sparsity. In a weight matrix,
this means the non-zero elements of each row should follow some regular
structure. \textbf{2) Neuron-level structured sparsity:} The surviving
neurons of each layer should also have some regular structures. In
a weight matrix, this means that the non-zero rows should also follow
some structure. \textbf{3) Flexibility:} The structure of the resulting
weight matrix should be sufficiently flexible so that we can achieve
a trade off between model complexity and predicting accuracy in different
scenarios. Fig. \ref{fig:Proposed-structure} illustrates an example
of a desired structure in the weight matrix. In each row, the significant
values tend to gather in groups. When translated into neural network
topology, this enables the advanced weight-level structure: The surviving
weights of each neuron tend to gather in groups. In each column, the
significant values also tend to gather in groups. When translated
into neural network topology, this enables the neuron-level structure:
The surviving neurons in each layer tend to gather in groups. Moreover,
the group size in each row and column can be tuned to achieve a trade
off between model complexity and predicting accuracy. Such proposed
multi-level structures enable the datapath to exploit the compression
to simplify the computation with very small control overheads. Specifically,
the advantages of our proposed multi-level structure are listed as
follows.
\begin{enumerate}
\item \emph{High decoding and computing parallelism:} As illustrated in
Fig. \ref{fig:pros and cons} (a), with the blocked structure, the
decoding can be performed in block wise such that all the weights
in the same block can be decoded simultaneously. Moreover, the regular
structure can bring more parallelism in computation. For example,
more efficient matrix tiling can be applied, where the zero blocks
can be skipped during matrix multiplication, as illustrated in Fig.
\ref{fig:pros and cons} (b). Also, the kernels can be compressed
into smaller size, such that the inputs corresponding to the zero
weights can be skipped \cite{zhang2019eager}.
\item \emph{Efficient neuron pruning:} As illustrated in Fig. \ref{fig:pros and cons}
(c), the structured surviving weights from the previous layer can
lead to a structured and compact neuron activation in the next layer,
which is beneficial for more efficient neuron pruning.
\item \emph{More efficient storage:} With the blocked structure, we can
simply record the location and size for each block rather than recording
the exact location for each weight. For a weight matrix with average
cluster size of $m\times m$ and number of clusters $P$, the coding
gain over traditional COO format is $\mathcal{O}\left(Pm^{2}\right)$.
\end{enumerate}
In this paper, we focus on promoting a multi-level group structure
in model compression. In our proposed structure, the surviving weights
connected to a neuron tend to gather in groups and the surviving neurons
also tend to gather in groups. Additionally, the group size is tunable
so that our proposed structure is flexible.
\begin{center}
\begin{figure*}[tb]
\begin{centering}
\includegraphics[width=0.85\textwidth]{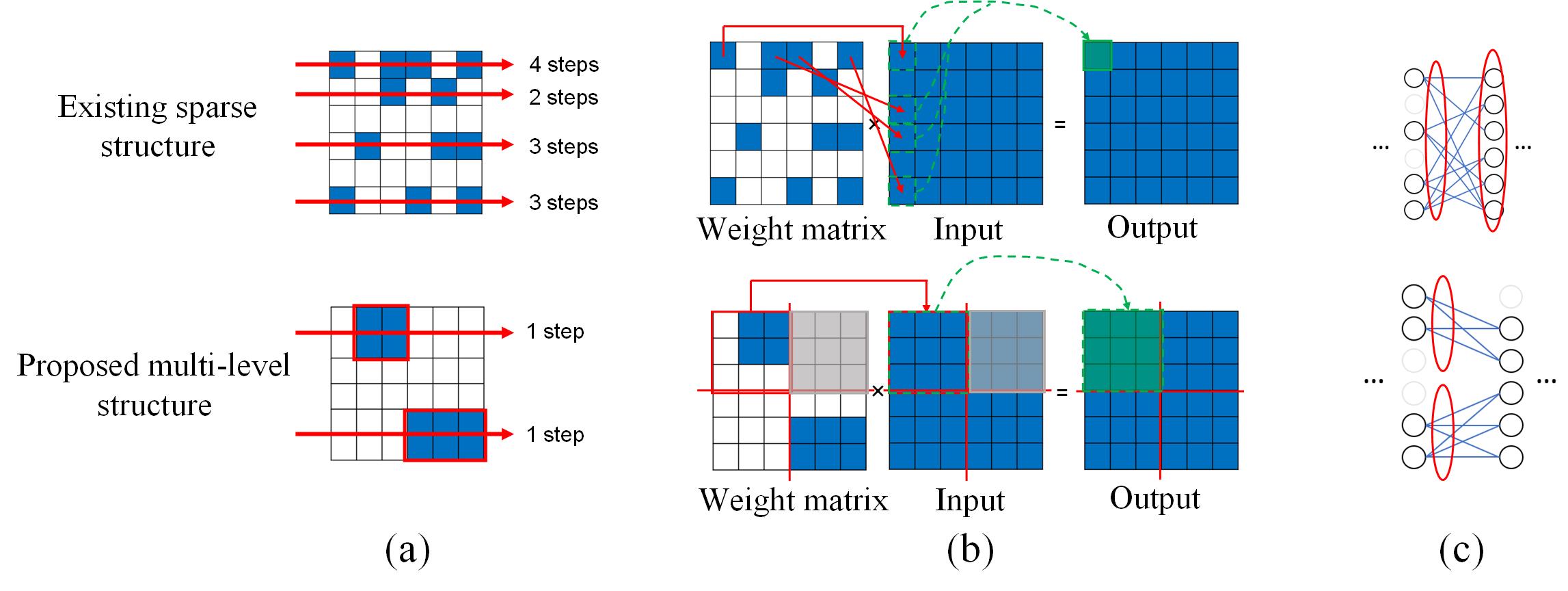}
\par\end{centering}
\caption{(a) Illustration of decoding parallelism. In existing sparse structures,
the decoding step for each row can be vastly different because of
the irregular structure. In our proposed structure, each block can
be decoded simultaneously because each block can be coded as a whole.
(b) Illustration of computing parallelism. In existing sparse structures,
the matrix-matrix multiplication has to be performed element-wise
because each non-zero weight is individually encoded. The processor
has to fetch individual weights and multiply them to individual input
elements to get the individual output element (green square). In our
proposed structure, parallel techinics for dense matrix such as matrix
tiling can be applied. Moreover, due to the clustered structure, the
zero block (gray block) can be skipped during computing the output
(green block). (c) Illustration of efficient neuron pruning. In existing
sparse structures, the surviving weights of each neuron are randomly
connected to the neurons in the next layer, which will randomly activate
neurons in the next layer. In our proposed structure, the surviving
weights of each neuron tend to activate the same neurons in the next
layer, which can lead to a structured and compact neuron activation
in the next layer.\label{fig:pros and cons}}
\end{figure*}
\par\end{center}

\subsection{Three-Layer Hierarchical Prior Model\label{subsec:Three-Layer-Hierarchical-Prior}}

The probability model of the sparse prior provides the foundation
of specific sparse structures. As previously mentioned, existing sparse
priors for model compression cannot promote the multi-level structure
in the weight matrix. To capture the multi-level structured sparsity,
we propose a three-layer hierarchical prior by combining the two-layer
prior and the support based prior \cite{Schniter2010}.

Let $w$ denote a weight in the neural network. For each weight $w$,
we introduce a support $s\in\left\{ 0,1\right\} $ to indicate whether
the weight $w$ is active $\left(s=1\right)$ or inactive $\left(s=0\right)$.
Specifically, let $\rho$ denote the precision for the weight $w$.
That is, $1/\rho$ is the variance of $w$. When $s=0$, the distribution
of $\rho$ is chosen to satisfy $\mathbb{E}\left[\rho\right]\gg1$,
so that the variance of the corresponding weight $w$ is very small.
When $s=1$, the distribution of $\rho$ is chosen to satisfy $\mathbb{E}\left[\rho\right]=\mathcal{O}\left(1\right)$,
so that $w$ has some probability to take significant values. Then,
the three-layer hierarchical prior (joint distribution of $\mathbf{w},\boldsymbol{\rho},\mathbf{s}$)
is given by
\begin{equation}
p\left(\mathbf{w},\boldsymbol{\rho},\mathbf{s}\right)=p\left(\mathbf{s}\right)p\left(\boldsymbol{\rho}|\mathbf{s}\right)p\left(\mathbf{w}|\boldsymbol{\rho}\right),\label{eq:3layer prior}
\end{equation}
 where $\mathbf{w}$ denotes all the weights in the neural network,
$\boldsymbol{\rho}$ denotes the corresponding precisions, and $\mathbf{s}$
denotes the corresponding supports. The distribution of each layer
is elaborated below.

\begin{figure}[tb]
\begin{centering}
\includegraphics[width=0.45\textwidth]{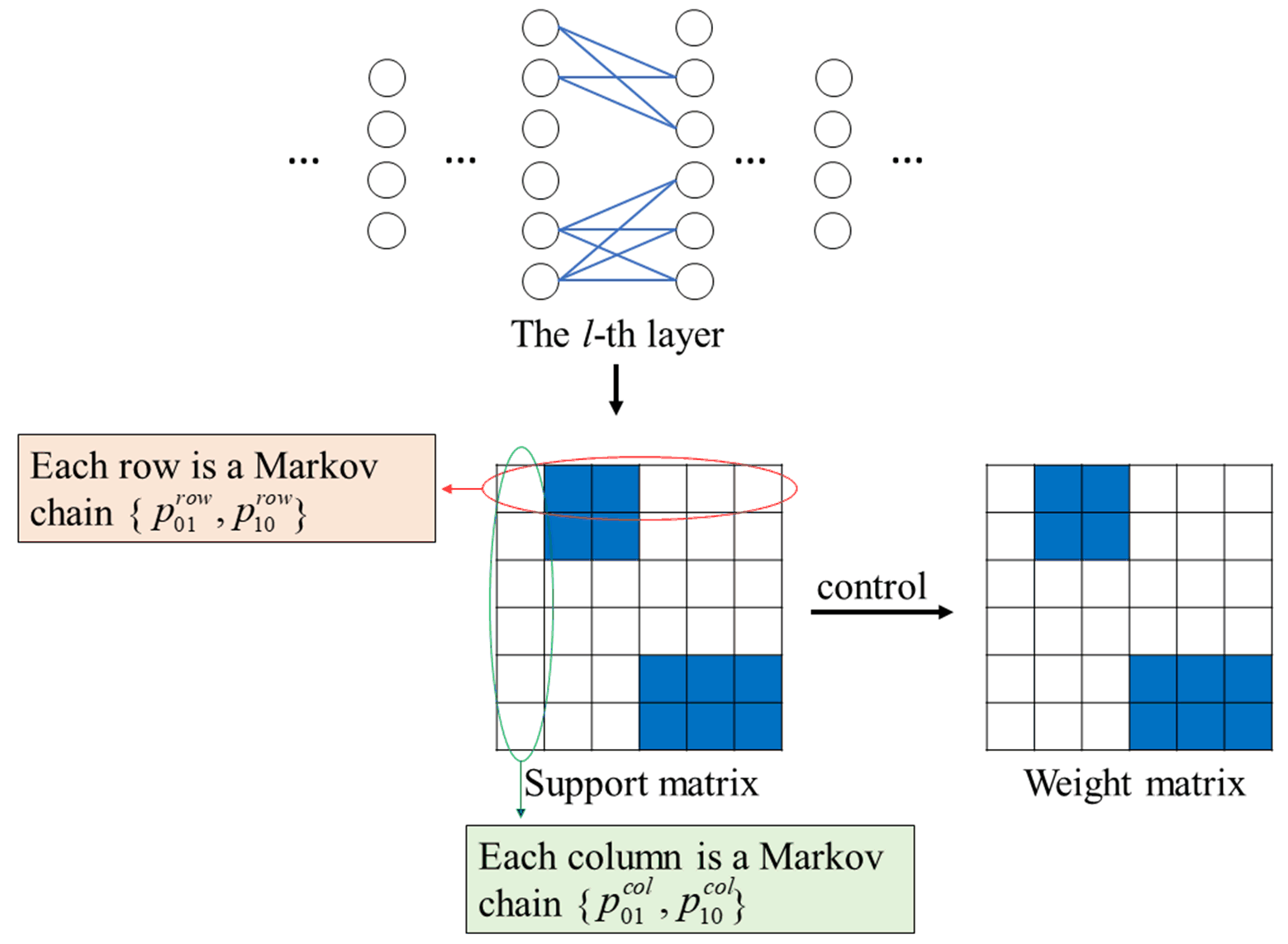}
\par\end{centering}
\caption{Illustration of the support prior for a $6\times6$ weight matrix.\label{fig:Illustration-of-support-prior}}
\end{figure}

\textbf{\emph{Probability Model for $p\left(\mathbf{s}\right)$:}}
$p\left(\mathbf{s}\right)$ can be decomposed into each layer by $p\left(\mathbf{s}\right)=\prod_{l=1}^{L}p\left(\mathbf{s}_{l}\right)$,
where $L$ is the total layer number of the neural network and $\mathbf{s}_{l}$
denotes the supports of the weights in the $l$-th layer. Now we focus
on the $l$-th layer. Suppose the dimension of the weight matrix is
$K\times M$, i.e., the layer has $K$ input neurons and $M$ output
neurons. The distribution $p\left(\mathbf{s}_{l}\right)$ of the supports
is used to capture the multi-level structure in this layer. Since
we focus on a specific layer now, we use $p\left(\mathbf{s}\right)$
to replace $p\left(\mathbf{s}_{l}\right)$ in the following discussion
for notation simplicity. In order to promote the aforementioned multi-level
structure, the active elements in each row of the weight matrix should
gather in clusters and the active elements in each column of the weight
matrix should also gather in clusters. Such a block structure can
be achieved by modeling the support matrix as a Markov random field
(MRF). Specifically, each row of the support matrix is modeled by
a Markov chain as
\begin{equation}
p\left(\mathbf{s}_{row}\right)=p\left(s_{row,1}\right)\prod_{m=1}^{M}p\left(s_{row,m+1}|s_{row,m}\right),
\end{equation}
 where $\mathbf{s}_{row}$ denotes the row vector of the support matrix,
and $s_{row,m}$ denotes the $m$-th element in $\mathbf{s}_{row}$.
The transition probability is given by $p\left(s_{row,m+1}=1|s_{row,m}=0\right)=p_{01}^{row}$
and $p\left(s_{row,m+1}=0|s_{row,m}=1\right)=p_{10}^{row}$. Generally,
a smaller $p_{01}^{row}$ leads to a larger average gap between two
clusters, and a smaller $p_{10}^{row}$ leads to a larger average
cluster size. Similarly, each column of the support matrix is also
modeled by a Markov chain as
\begin{equation}
p\left(\mathbf{s}_{col}\right)=p\left(s_{col,1}\right)\prod_{k=1}^{K}p\left(s_{col,k+1}|s_{col,k}\right),
\end{equation}
 where $\mathbf{s}_{col}$ denotes the column vector of the support
matrix and $s_{col,k}$ denotes the $n$-th element in $\mathbf{s}_{col}$.
The transition probability is given by $p\left(s_{col,k+1}=1|s_{col,k}=0\right)=p_{01}^{col}$
and $p\left(s_{col,k+1}=0|s_{col,k}=1\right)=p_{10}^{col}$. Such
an MRF model is also known as a 2D Ising model. An illustration of
the MRF prior is given in Fig. \ref{fig:Illustration-of-support-prior}.
Note that $p\left(\mathbf{s}\right)$ can also be modeled with other
types of priors, such as Markov tree prior and hidden Markov prior,
to promote other structures.

\textbf{\emph{Probability Model for $p\left(\boldsymbol{\rho}|\mathbf{s}\right)$:}}
The conditional probability $p\left(\boldsymbol{\rho}|\mathbf{s}\right)$
is given by 
\begin{equation}
p\left(\boldsymbol{\rho}|\mathbf{s}\right)=\prod_{n=1}^{N}\left(\Gamma\left(\rho_{n};a_{n},b_{n}\right)\right)^{s_{n}}\left(\Gamma\left(\rho_{n};\overline{a}_{n},\overline{b}_{n}\right)\right)^{1-s_{n}},
\end{equation}
 where $N$ is the total number of weights in the neural network.
$\rho_{n}$ takes two different Gamma distributions according to the
value of $s_{n}$. When $s_{n}=1$, the corresponding weight $w_{n}$
is active. In this case, the shape parameter $a_{n}$ and rate parameter
$b_{n}$ should satisfy that $\frac{a_{n}}{b_{n}}=\mathbb{E}\left[\rho_{n}\right]=\mathcal{O}\left(1\right)$
such that the variance of $w_{n}$ is $\mathcal{O}\left(1\right)$.
When $s_{n}=0$, the corresponding weight $w_{n}$ is inactive. In
this case, the shape parameter $\overline{a}_{n}$ and rate parameter
$\overline{b}_{n}$ should satisfy that $\frac{\overline{a}_{n}}{\overline{b}_{n}}=\mathbb{E}\left[\rho_{n}\right]\gg1$
such that the variance of $w_{n}$ is very close to zero. The motivation
for choosing Gamma distribution is that Gamma distribution is conjugate
to Gaussian distribution. Thus, it can leads to a closed form solution
in Bayesian inference \cite{Liu2020}.

\textbf{\emph{Probability Model for $p\left(\mathbf{w}|\boldsymbol{\rho}\right)$:}}\emph{
}The conditional probability $p\left(\mathbf{w}|\boldsymbol{\rho}\right)$
is given by 
\begin{equation}
p\left(\mathbf{w}|\boldsymbol{\rho}\right)=\prod_{n=1}^{N}p\left(w_{n}|\rho_{n}\right),
\end{equation}
 where $p\left(w_{n}|\rho_{n}\right)$ is assumed to be a Gaussian
distribution:
\begin{equation}
p\left(w_{n}|\rho_{n}\right)=\mathcal{N}\left(w_{n}|0,\frac{1}{\rho_{n}}\right).
\end{equation}
 Although the choice of $p\left(w_{n}|\rho_{n}\right)$ is not restricted
to Gaussian distribution \cite{Louizos2017}, the motivation for choosing
Gaussian is still reasonable. First, according to existing simulations,
the compression performance is not so sensitive to the distribution
type of $p\left(w_{n}|\rho_{n}\right)$ \cite{Louizos2017,Molchanov2017}.
Moreover, assuming $p\left(w_{n}|\rho_{n}\right)$ to be a Gaussian
distribution also contributes to easier optimization in Bayesian inference,
as shown in Section \ref{sec:turbo-vbi-algorithm}.

\emph{Remark 1. (Comparison with existing sparse priors)} One of the
main differences between our work and the existing works \cite{Molchanov2017,Louizos2017}
is the design of the prior. Because of the MRF support layer, our
proposed three-layer prior is more general and can capture more regular
structures such as the desired multi-level structure. Such a multi-level
structure in the weight matrix cannot be modeled by the existing priors
in \cite{Molchanov2017} and \cite{Louizos2017} but our proposed
prior can easily support the sparse structures in them. In our work,
if the prior $p\left(\mathbf{w},\boldsymbol{\rho},\mathbf{s}\right)$
reduces to one layer $p\left(\mathbf{w}\right)$ and takes a Laplace
distribution, it is equivalent to a Lasso regularization. In this
way, random sparsity can be realized. If the prior $p\left(\mathbf{w},\boldsymbol{\rho},\mathbf{s}\right)$
reduces to one layer $p\left(\mathbf{w}\right)$ and takes an improper
log-scale uniform distribution, it is exactly the problem of variational
dropout \cite{Molchanov2017}. In this way, random sparsity can be
realized. If the prior $p\left(\mathbf{w},\boldsymbol{\rho},\mathbf{s}\right)$
reduces to a two-layer $p\left(\mathbf{w}|\boldsymbol{\rho}\right)$,
and takes the group improper log-uniform distribution or the group
horseshoe distribution, it is exactly the problem in \cite{Louizos2017}.
In this way, group sparsity can be realized. Moreover, since our proposed
prior is more complicated, it also leads to a different algorithm
design in Section \ref{sec:turbo-vbi-algorithm}.

\section{Bayesian Model Compression Formulation\label{sec:problem-formulation}}

Bayesian model compression handles the model compression problem from
a Bayesian perspective. In Bayesian model compression, the weights
follow a prior distribution, and our primary goal is to find the posterior
distribution conditioned on the dataset. In the following, we elaborate
on the neural network model and the Bayesian training of the DNN.

\subsection{Neural Network Model}

\begin{figure}[tb]
\subfloat[Illustration of $NN\left(x,\mathbf{w}\right)$ in MLP.\label{fig:Illustration in mlp}]{\begin{centering}
\includegraphics[width=0.45\textwidth]{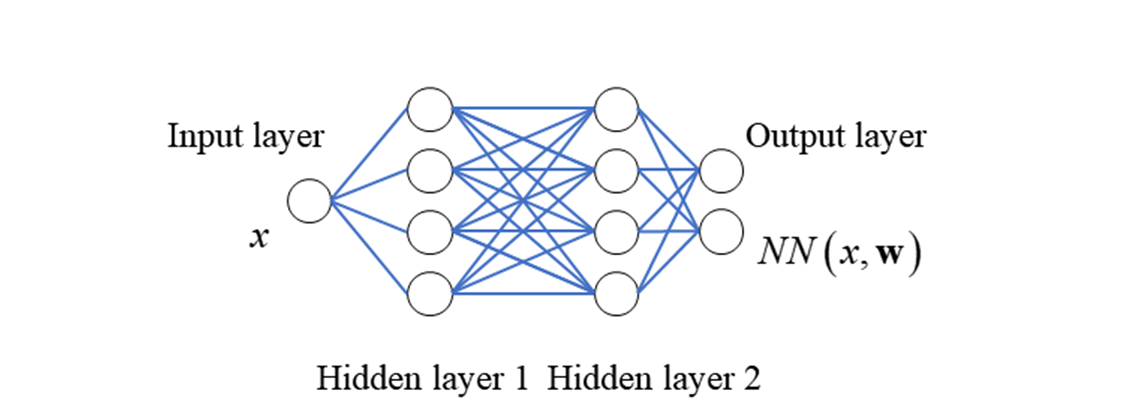}
\par\end{centering}
}

\subfloat[Illustration of $NN\left(x,\mathbf{w}\right)$ in CNN.\label{fig:Illustration in cnn}]{\begin{centering}
\includegraphics[width=0.45\textwidth]{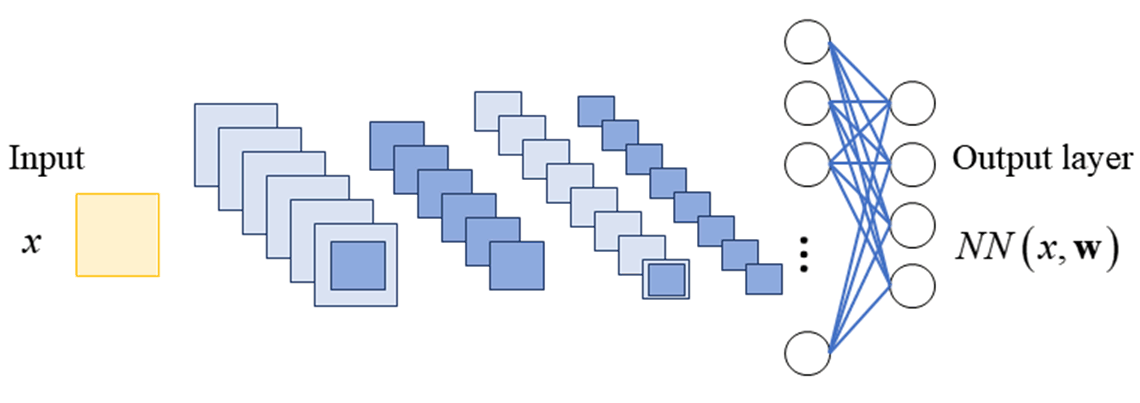}
\par\end{centering}
}

\caption{Illustration of $NN\left(x,\mathbf{w}\right)$ in different neural
networks.}
\end{figure}

We first introduce the neural network model. Let $\mathcal{D}=\left\{ \left(x_{1},y_{1}\right),\left(x_{2},y_{2}\right),...,\left(x_{D},y_{D}\right)\right\} $
denote the dataset, which contains $D$ pairs of training input $\left(x_{d}\right)$
and training output $\left(y_{d}\right)$. Let $NN\left(x,\mathbf{w}\right)$
denote the output of the neural network with input $x$ and weights
$\mathbf{w}$, where $NN\left(\cdot\right)$ is the neural network
function. Note that $NN\left(x,\mathbf{w}\right)$ is a generic neural
network model, which can embrace various commonly used structures,
some examples are listed as follows.
\begin{itemize}
\item \textbf{Multi-Layer Perceptron (MLP): }MLP is a representative class
of feedforward neural network, which consists of an input layer, output
layer and hidden layers. As illustrated in Fig. \ref{fig:Illustration in mlp},
we can use $NN\left(x,\mathbf{w}\right)$ to represent the feedforward
calculation in an MLP.
\item \textbf{Convolutional Neural Network (CNN): }A CNN consists of convolutional
kernels and dense layers. As illustrated in Fig. \ref{fig:Illustration in cnn},
we can use $NN\left(x,\mathbf{w}\right)$ to represent the complicated
calculation in a CNN.
\end{itemize}

Since the data points from the training set are generally assumed
to be independent, we can use the following stochastic model to describe
the observations from a general neural network:

\begin{equation}
\mathbf{y}\sim p\left(\mathcal{D}|\mathbf{w}\right)=\prod_{d=1}^{D}p\left(y_{d}|x_{d},\mathbf{w}\right),\label{eq:bayesian nn model}
\end{equation}
 where the likelihood $p\left(y_{d}|x_{d},\mathbf{w}\right)$ can
be different for regression and classification tasks \cite{Jospin2020}.
In regression tasks, it can be modeled as a Gaussian distribution:
\begin{equation}
p\left(y_{d}|x_{d},\mathbf{w}\right)=\mathcal{N}\left(NN\left(x_{d},\mathbf{w}\right),\sigma_{d}^{2}\right),\label{eq:bayesian nn model 2}
\end{equation}
 where $\sigma_{d}^{2}$ is the noise variance in training data, while
in classification tasks, it can be modeled as
\begin{equation}
p\left(y_{d}|x_{d},\mathbf{w}\right)=\exp\left(-G\left(y_{d}|x_{d},\mathbf{w}\right)\right),
\end{equation}
 where $G\left(y_{d}|x_{d},\mathbf{w}\right)$ is the cross-entropy
error function.

\subsection{Bayesian Training of DNNs}

Bayesian training of a DNN is actually calculating the posterior distribution
$p\left(\mathbf{w},\boldsymbol{\rho},\mathbf{s}|\mathcal{D}\right)$
based on the prior distribution $p\left(\mathbf{w},\boldsymbol{\rho},\mathbf{s}\right)$
and the neural network stochastic model \eqref{eq:bayesian nn model}.
In our research, after we obtain the posterior $p\left(\mathbf{w},\boldsymbol{\rho},\mathbf{s}|\mathcal{D}\right)$,
we use the maximum a posteriori (MAP) estimate of $\mathbf{w},\boldsymbol{\rho}$
and $\mathbf{s}$ as a deterministic estimate of the weights, precision
and support, respectively:
\begin{equation}
\left(\mathbf{w},\boldsymbol{\rho},\mathbf{s}\right)^{*}=\arg\max_{\mathbf{w},\boldsymbol{\rho},\mathbf{s}}p\left(\mathbf{w},\boldsymbol{\rho},\mathbf{s}|\mathcal{D}\right).
\end{equation}

According to Bayesian rule, the posterior distribution of $\mathbf{w},\boldsymbol{\rho}$
and $\mathbf{s}$ can be derived as 
\begin{equation}
p\left(\mathbf{w},\boldsymbol{\rho},\mathbf{s}|\mathcal{D}\right)=\frac{p\left(\mathcal{D}|\mathbf{w},\boldsymbol{\rho},\mathbf{s}\right)p\left(\mathbf{w},\boldsymbol{\rho},\mathbf{s}\right)}{p\left(\mathcal{D}\right)},\label{eq:posterior}
\end{equation}
 where $p\left(\mathcal{D}|\mathbf{w},\boldsymbol{\rho},\mathbf{s}\right)$
is the likelihood term. Based on \eqref{eq:bayesian nn model} and
\eqref{eq:bayesian nn model 2}, it is given by:
\begin{equation}
p\left(\mathcal{D}|\mathbf{w},\boldsymbol{\rho},\mathbf{s}\right)=\prod_{d=1}^{D}p\left(y_{d}|x_{d},\mathbf{w},\boldsymbol{\rho},\mathbf{s}\right),
\end{equation}
\begin{equation}
p\left(y_{d}|x_{d},\mathbf{w},\boldsymbol{\rho},\mathbf{s}\right)=\mathcal{N}\left(NN\left(x_{d},\mathbf{w},\boldsymbol{\rho},\mathbf{s}\right),\sigma_{d}^{2}\right).
\end{equation}
 Equation \eqref{eq:posterior} mainly contains two terms: 1) Likelihood
$p\left(\mathcal{D}|\mathbf{w},\boldsymbol{\rho},\mathbf{s}\right)$,
which is related to the expression of dataset. 2) Prior $p\left(\mathbf{w},\boldsymbol{\rho},\mathbf{s}\right)$,
which is related to the sparse structure we want to promote. Intuitively,
by Bayesian training of the DNN, we are actually simultaneously 1)
tuning the weights to express the dataset and 2) tuning the weights
to promote the structure captured in the prior.

However, directly computing the posterior according to \eqref{eq:posterior}
involves several challenges, summarized as follows.
\begin{itemize}
\item \textbf{Intractable multidimensional integral:} The calculation of
the exact posterior $p\left(\mathbf{w},\boldsymbol{\rho},\mathbf{s}|\mathcal{D}\right)$
involves calculating the so-called evidence term $p\left(\mathcal{D}\right)$.
The calculation of $p\left(\mathcal{D}\right)=\sum_{\mathbf{s}}\int\int p\left(\mathcal{D}|\mathbf{w},\boldsymbol{\rho},\mathbf{s}\right)p\left(\mathbf{w},\boldsymbol{\rho},\mathbf{s}\right)d\mathbf{w}d\boldsymbol{\rho}$
is usually intractable as the likelihood term can be very complicated
\cite{Jospin2020}.
\item \textbf{Complicated prior term:} Since the prior term $p\left(\mathbf{w},\boldsymbol{\rho},\mathbf{s}\right)$
contains an extra support layer $p\left(\mathbf{s}\right)$, the KL-divergence
between the approximate distribution and the posterior distribution
is difficult to calculate or approximate. Traditional VBI method cannot
be directly applied to achieve an approximate posterior distribution.
\end{itemize}
In the next section, we propose a Turbo-VBI based model compression
method, which approximately calculates the marginal posterior distribution
of $\mathbf{w},\boldsymbol{\rho}$ and $\mathbf{s}$.

\section{Turbo-VBI Algorithm for Model Compression\label{sec:turbo-vbi-algorithm}}

\begin{figure}[tb]
\begin{centering}
\includegraphics[width=0.45\textwidth]{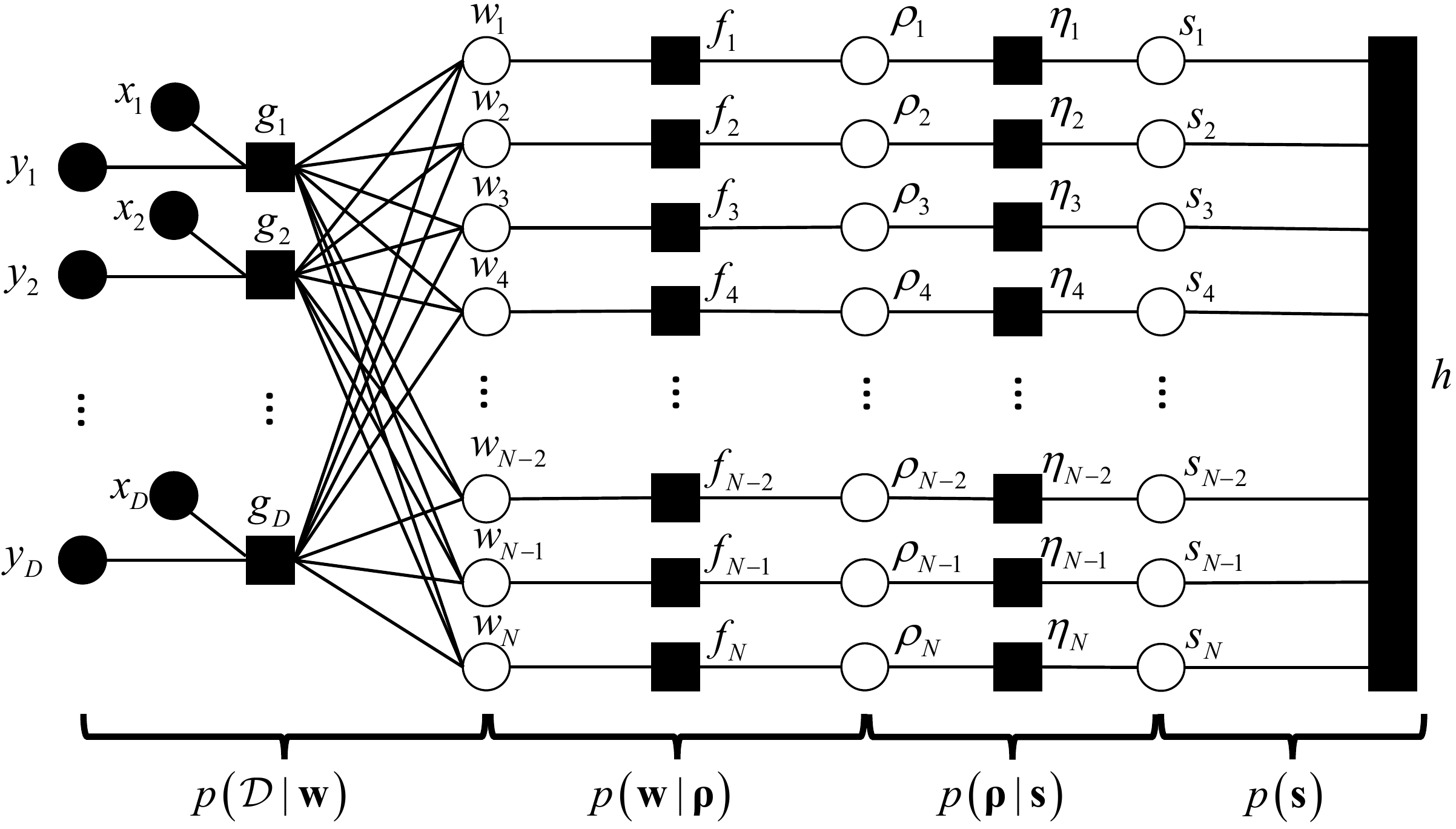}
\par\end{centering}
\caption{Factor graph of the joint distribution $p\left(\mathbf{w},\boldsymbol{\rho},\mathbf{s},\mathcal{D}\right)$.\label{fig:Factor-graph}}
\end{figure}

To handle the aforementioned two challenges, we propose a Turbo-VBI
\cite{Liu2020} based model compression algorithm. The basic idea
of the proposed Turbo-VBI algorithm is to approximate the intractable
posterior $p\left(\mathbf{w},\boldsymbol{\rho},\mathbf{s}|\mathcal{D}\right)$
with a variational distribution $q\left(\mathbf{w},\boldsymbol{\rho},\mathbf{s}\right)$.
The factor graph of the joint distribution $p\left(\mathbf{w},\boldsymbol{\rho},\mathbf{s},\mathcal{D}\right)$
is illustrated in Fig. \ref{fig:Factor-graph}, where the variable
nodes are denoted by white circles and the factor nodes are denoted
by black squares. Specifically, the factor graph is derived based
on the factorization 
\begin{equation}
p\left(\mathbf{w},\boldsymbol{\rho},\mathbf{s},\mathcal{D}\right)=p\left(\mathcal{D}|\mathbf{w}\right)p\left(\mathbf{w}|\boldsymbol{\rho}\right)p\left(\boldsymbol{\rho}|\mathbf{s}\right)p\left(\mathbf{s}\right).
\end{equation}
As illustrated in Fig. \ref{fig:Factor-graph}, the variable nodes
are $\mathbf{w},\boldsymbol{\rho}$ and $\mathbf{s}$, $g$ denotes
the likelihood function, $f$ denotes the prior distribution $p\left(w_{n}|\rho_{n}\right)$
for weights $\mathbf{w}$, $\eta$ denotes the prior distribution
$p\left(\rho_{n}|s_{n}\right)$ for precision $\boldsymbol{\rho}$,
and $h$ denotes the joint prior distribution $p\left(\mathbf{s}\right)$
for support $\mathbf{s}$. The detailed expression of each factor
node is listed in Table \ref{tab:Detailed-expression-of-function-node}.

\begin{table*}[tb]
\begin{centering}
\caption{Detailed expression of each factor node.\label{tab:Detailed-expression-of-function-node}}
\par\end{centering}
\centering{}%
\begin{tabular}{|c|c|c|}
\hline 
Factor node & Distribution & Function\tabularnewline
\hline 
$g_{d}\left(x_{d},y_{d},\mathbf{w}\right)$ & $p\left(y_{d}|x_{d},\mathbf{w}\right)$ & $\mathcal{N}\left(NN\left(x_{d},\mathbf{w}\right),\sigma_{d}^{2}\right)$\tabularnewline
\hline 
$f_{n}\left(w_{n},\rho_{n}\right)$ & $p\left(w_{n}|\rho_{n}\right)$ & $\mathcal{N}\left(w_{n}|0,\frac{1}{\rho_{n}}\right)$\tabularnewline
\hline 
$\eta_{n}\left(\rho_{n},s_{n}\right)$ & $p\left(\rho_{n}|s_{n}\right)$ & $\left(\Gamma\left(\rho_{n};a_{n},b_{n}\right)\right)^{s_{n}}\left(\Gamma\left(\rho_{n};\overline{a}_{n},\overline{b}_{n}\right)\right)^{1-s_{n}}$\tabularnewline
\hline 
$h$ & $p\left(\mathbf{s}\right)$ & MRF prior. $p\left(s_{row,m+1}|s_{row,m}\right),p\left(s_{col,k+1}|s_{col,k}\right)$\tabularnewline
\hline 
\end{tabular}
\end{table*}

\subsection{Top Level Modules of the Turbo-VBI Based Model Compression Algorithm}

\begin{figure}[tb]
\begin{centering}
\includegraphics[width=0.45\textwidth]{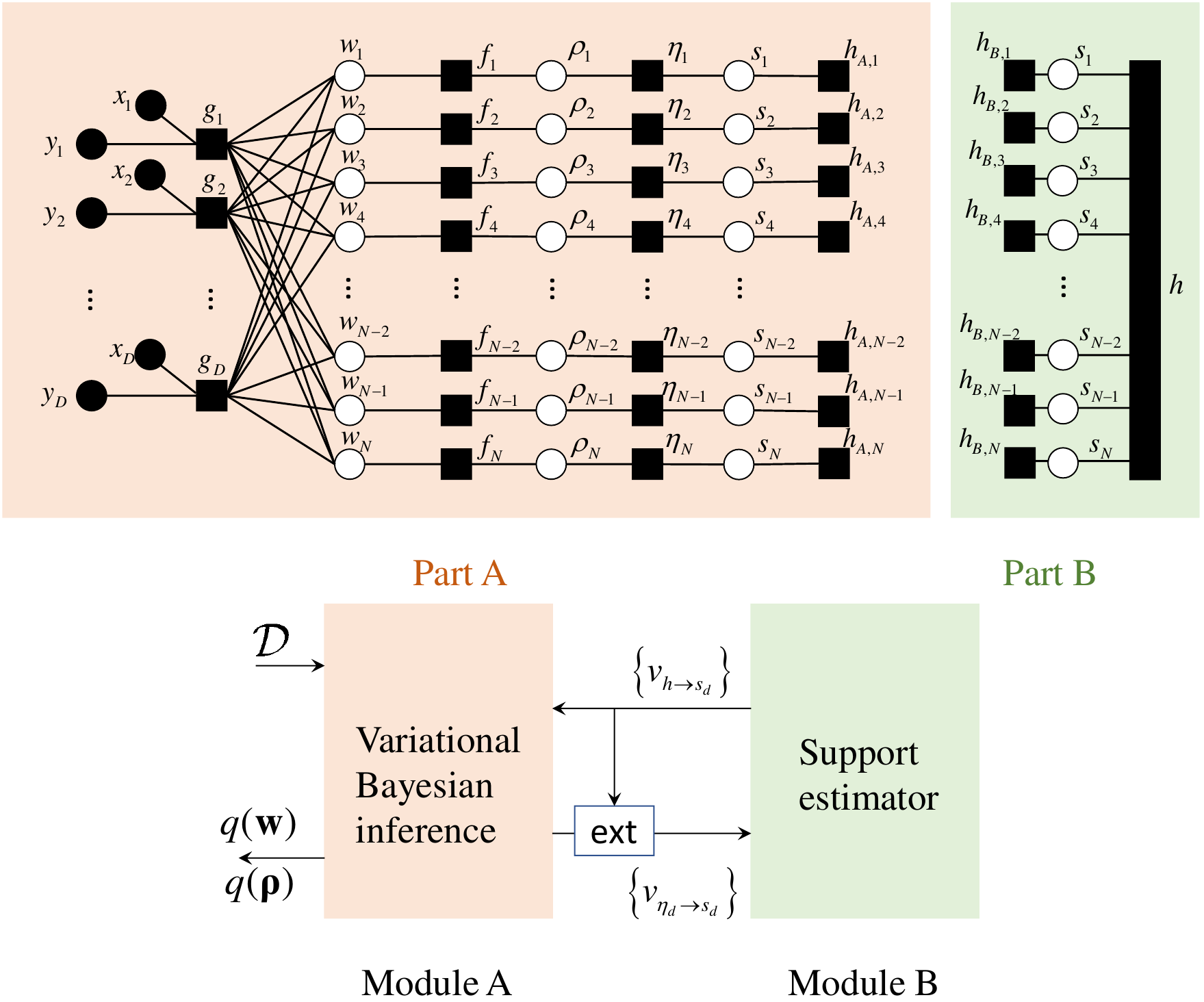}
\par\end{centering}
\centering{}\caption{Illustration of the two modules in the Turbo-VBI algorithm.\label{fig:Illustration-of-2module}}
\end{figure}

Since the factor graph in Fig. \ref{fig:Factor-graph} has loops,
directly applying the message passing algorithm usually cannot achieve
a good performance. In addition, since the probability model is more
complicated than the existing ones in \cite{Molchanov2017} and \cite{Louizos2017},
it is difficult to directly apply the VBI algorithm. Thus, we consider
separating the complicated probability model in Fig. \ref{fig:Factor-graph}
into two parts, and perform Bayesian inference respectively, as illustrated
in Fig. \ref{fig:Illustration-of-2module}. Specifically, we follow
the turbo framework and divide the factor graph into two parts. One
is the support part (Part B), which contains the prior information.
The other one is the remaining part (Part A). Correspondingly, the
proposed Turbo-VBI algorithm is also divided into two modules such
that each module performs Bayesian inference on its corresponding
part respectively. Module A and Module B also need to exchange messages.
Specifically, the output message $v_{h\rightarrow s_{n}}$ of Module
B refers to the message from factor node $h_{A,n}$ to variable node
$s_{n}$ in Part A, and is equivalent to the message from variable
node $s_{n}$ to factor node $h_{B,n}$ in Part B, which refers to
the conditional marginal probability $p\left(s_{n}|\boldsymbol{v}_{\boldsymbol{\eta}\rightarrow\mathbf{s}}\right)$.
It is calculated by sum-product message passing (SPMP) on part B,
and acts as the input of Module A to facilitate the VBI on Part A.
The output message $v_{\eta_{n}\rightarrow s_{n}}$ of Module A refers
to the posterior probability subtracting the output of Module B: $\frac{q\left(s_{n}\right)}{v_{h\rightarrow s_{n}}}$,
and acts as the input of Module B to facilitate SPMP on Part B.

Specifically, the probability model for the prior distribution in
Part A is assumed as
\begin{equation}
\hat{p}\left(\mathbf{w},\boldsymbol{\rho},\mathbf{s}\right)=\hat{p}\left(\mathbf{s}\right)p\left(\boldsymbol{\rho}|\mathbf{s}\right)p\left(\mathbf{w}|\boldsymbol{\rho}\right),\label{eq:new prior}
\end{equation}
 where 
\[
\hat{p}\left(\mathbf{s}\right)=\prod_{n=1}^{N}\left(\pi_{n}\right)^{s_{n}}\left(1-\pi_{n}\right)^{1-s_{n}}
\]
 is a new prior for support $\mathbf{s}$. $\pi_{n}$ is the probability
that $s_{n}=1$, which is defined as:
\begin{equation}
\pi_{n}=p\left(s_{n}=1\right)=\frac{v_{h\rightarrow s_{n}}\left(1\right)}{v_{h\rightarrow s_{n}}\left(1\right)+v_{h\rightarrow s_{n}}\left(0\right)}.
\end{equation}
 Note that the only difference between the new prior in \eqref{eq:new prior}
and that in \eqref{eq:3layer prior} is that the complicated $p\left(\mathbf{s}\right)$
is replaced by a simpler distribution $\hat{p}\left(\mathbf{s}\right)$
with independent entries. The correlations among the supports are
separated into Part B. Then, based on the new prior $\hat{p}\left(\mathbf{w},\boldsymbol{\rho},\mathbf{s}\right)$,
Module A performs a VBI algorithm. In the VBI algorithm, variational
distributions $q\left(\mathbf{w}\right)$, $q\left(\boldsymbol{\rho}\right)$
and $q\left(\mathbf{s}\right)$ are used to approximate the posterior.
After that, the approximate posterior $q\left(\mathbf{s}\right)$
is delivered from Module A back into Module B. The delivered message
$v_{\eta_{n}\rightarrow s_{n}}$ is defined as 
\begin{equation}
v_{\eta_{n}\rightarrow s_{n}}=\frac{q\left(s_{n}\right)}{v_{h\rightarrow s_{n}}},\label{eq:messgae A to B}
\end{equation}
 according to the sumproduct rule.

With the input message $v_{\eta_{n}\rightarrow s_{n}}$, Module B
further performs SPMP over Part B to exploit the structured sparsity,
which is contained in the prior distribution $p\left(\mathbf{s}\right).$
Specifically, in Part B, the factor nodes
\begin{equation}
h_{B,n}=v_{\eta_{n}\rightarrow s_{n}},n=1,...,N\label{eq:hbn}
\end{equation}
 carry the information of the variational posterior distribution $q\left(\mathbf{s}\right)$,
and the factor node $h$ carries the structured prior information
of $p\left(\mathbf{s}\right)$. By performing SPMP over Part B, the
message $v_{h\rightarrow s_{n}}$ can be calculated and then delivered
to Module A. These two modules exchange messages iteratively until
convergence or the maximum iteration number is exceeded. After this,
the final outputs $q\left(\mathbf{w}\right)$, $q\left(\boldsymbol{\rho}\right)$
and $q\left(\mathbf{s}\right)$ of Module A are the approximate posterior
distribution for $\mathbf{w}$, $\boldsymbol{\rho},$ and $\mathbf{s}$,
respectively. Note that although the SPMP is not guaranteed to converge
on Part B, because our $p\left(\mathbf{s}\right)$ is an MRF prior,
the desired structure is usually well promoted in the final output
of Module A, as illustrated in the simulation results. This is because
SPMP can achieve good results on 2-D Ising model and there have been
many solid works concerning SPMP on loopy graphs \cite{Murphy2013}.
And our proposed VBI compression algorithm in Module A can achieve
good enough performance to capture the structure. In addition, since
the design of $p\left(\mathbf{s}\right)$ is flexible, one can model
it with Markov chain priors or Markov tree priors. In this case, the
convergence is guaranteed. In the following, we elaborate the two
modules in details.

\subsection{Sparse VBI Estimator (Module A)}

In Module A, we want to use a tractable variational distribution $q\left(\mathbf{w},\boldsymbol{\rho},\mathbf{s}\right)$
to approximate the intractable posterior distribution $\hat{p}\left(\mathbf{w},\boldsymbol{\rho},\mathbf{s}|\mathcal{D}\right)$
under the prior $\hat{p}\left(\mathbf{w},\boldsymbol{\rho},\mathbf{s}\right)$.
The quality of this approximation is measured by the Kullback-Leibler
divergence (KL divergence):
\begin{multline}
D_{KL}\left(q\left(\mathbf{w},\boldsymbol{\rho},\mathbf{s}\right)||\hat{p}\left(\mathbf{w},\boldsymbol{\rho},\mathbf{s}|\mathcal{D}\right)\right)\\
=\sum_{\mathbf{s}}\int\int q\left(\mathbf{w},\boldsymbol{\rho},\mathbf{s}\right)\ln\frac{q\left(\mathbf{w},\boldsymbol{\rho},\mathbf{s}\right)}{\hat{p}\left(\mathbf{w},\boldsymbol{\rho},\mathbf{s}|\mathcal{D}\right)}d\mathbf{w}d\boldsymbol{\rho}.
\end{multline}
 Thus, the goal is to find the optimal variational distribution $q\left(\mathbf{w},\boldsymbol{\rho},\mathbf{s}\right)$
that minimizes this KL divergence. However, the KL divergence still
involves calculating the intractable posterior $\hat{p}\left(\mathbf{w},\boldsymbol{\rho},\mathbf{s}|\mathcal{D}\right)$.
In order to overcome this challenge, the minimization of KL divergence
is usually transformed into a minimization of the negative evidence
lower bound (ELBO), subject to a factorized form constraint \cite{Tzikas2008}.

\textbf{Sparse VBI Problem:}

\begin{equation}
\begin{aligned}\min_{q\left(\mathbf{w},\boldsymbol{\rho},\mathbf{s}\right)} & -ELBO,\\
s.t. & q\left(\mathbf{w},\boldsymbol{\rho},\mathbf{s}\right)=\prod_{n=1}^{N}q\left(w_{n}\right)q\left(\rho_{n}\right)q\left(s_{n}\right),
\end{aligned}
\label{eq:min_neg_ELBO}
\end{equation}
 where $ELBO=\sum_{\mathbf{s}}\int\int q\left(\mathbf{w},\boldsymbol{\rho},\mathbf{s}\right)\ln p\left(\mathcal{D}|\mathbf{w},\boldsymbol{\rho},\mathbf{s}\right)d\mathbf{w}d\boldsymbol{\rho}-D_{KL}\left(q\left(\mathbf{w},\boldsymbol{\rho},\mathbf{s}\right)||\hat{p}\left(\mathbf{w},\boldsymbol{\rho},\mathbf{s}\right)\right).$
The constraint means all individual variables $w,\rho$ and $s$ are
assumed to be independent. Such an assumption is known as the mean
field assumption and is widely used in VBI methods \cite{Louizos2017,Tzikas2008}.

The problem in \eqref{eq:min_neg_ELBO} is non-convex and generally
we can only achieve a stationary solution $q^{*}\left(\mathbf{w},\boldsymbol{\rho},\mathbf{s}\right)$.
A stationary solution can be achieved by applying a block coordinate
descent (BCD) algorithm to the problem in \eqref{eq:min_neg_ELBO},
as will be proved in Lemma 1. According to the idea of the BCD algorithm,
to solve \eqref{eq:min_neg_ELBO} is equivalent to iteratively solving
the following three subproblems.

\textbf{\emph{Subproblem 1 (update of $\boldsymbol{\rho}$):}}\emph{
}
\begin{equation}
\begin{aligned}\min_{q\left(\boldsymbol{\rho}\right)} & D_{KL}\left(q\left(\boldsymbol{\rho}\right)||\widetilde{p}_{\boldsymbol{\rho}}\right),\\
s.t. & q\left(\boldsymbol{\rho}\right)=\prod_{n=1}^{N}q\left(\rho_{n}\right),
\end{aligned}
\label{eq:subproblem 1}
\end{equation}
 where $\ln\widetilde{p}_{\boldsymbol{\rho}}=\left\langle \ln\hat{p}\left(\mathbf{w},\boldsymbol{\rho},\mathbf{s},\mathcal{D}\right)\right\rangle _{q\left(\mathbf{s}\right)q\left(\mathbf{w}\right)}.$

\textbf{\emph{Subproblem 2 (update of $\mathbf{s}$):}}\emph{ }
\begin{equation}
\begin{aligned}\min_{q\left(\mathbf{s}\right)} & D_{KL}\left(q\left(\boldsymbol{s}\right)||\widetilde{p}_{\mathbf{s}}\right),\\
s.t. & q\left(\mathbf{s}\right)=\prod_{n=1}^{N}q\left(s_{n}\right),
\end{aligned}
\label{eq:subproblem 2}
\end{equation}
 where $\ln\widetilde{p}_{\mathbf{s}}=\left\langle \ln\hat{p}\left(\mathbf{w},\boldsymbol{\rho},\mathbf{s},\mathcal{D}\right)\right\rangle _{q\left(\boldsymbol{\rho}\right)q\left(\mathbf{w}\right)}.$

\textbf{\emph{Subproblem 3 (update of $\mathbf{w}$):}} 
\begin{equation}
\begin{aligned}\min_{q\left(\mathbf{w}\right)} & D_{KL}\left(q\left(\mathbf{w}\right)||\widetilde{p}_{\mathbf{w}}\right),\\
s.t. & q\left(\mathbf{w}\right)=\prod_{n=1}^{N}q\left(w_{n}\right),
\end{aligned}
\label{eq:subproblem 3}
\end{equation}
 where $\ln\widetilde{p}_{\mathbf{w}}=\left\langle \ln\hat{p}\left(\mathbf{w},\boldsymbol{\rho},\mathbf{s},\mathcal{D}\right)\right\rangle _{q\left(\boldsymbol{\rho}\right)q\left(\mathbf{s}\right)}$
and $\left\langle f\left(x\right)\right\rangle _{q\left(x\right)}=\int f\left(x\right)q\left(x\right)dx.$
The detailed derivations of the three subproblems are provided in
the Appendix A. In the following, we elaborate on solving the three
subproblems respectively.

\emph{1) Update of }\textbf{\emph{$q\left(\boldsymbol{\rho}\right)$:}}\textbf{
}In subproblem 1, obviously, the optimal solution $q^{\star}\left(\boldsymbol{\rho}\right)$
is achieved when $q^{\star}\left(\boldsymbol{\rho}\right)=\widetilde{p}_{\boldsymbol{\rho}}$.
In this case, $q^{\star}\left(\boldsymbol{\rho}\right)$ can be derived
as 
\begin{equation}
q^{\star}\left(\boldsymbol{\rho}\right)=\prod_{n=1}^{N}\Gamma\left(\rho_{n};\widetilde{a}_{n},\widetilde{b}_{n}\right),
\end{equation}
 where $\widetilde{a}_{n}$ and $\widetilde{b}_{n}$ are given by:
\begin{equation}
\begin{aligned}\widetilde{a}_{n} & =\left\langle s_{n}\right\rangle a_{n}+\left\langle 1-s_{n}\right\rangle \overline{a}_{n}+1\\
 & =\widetilde{\pi}_{n}a_{n}+\left(1-\widetilde{\pi}_{n}\right)\overline{a}_{n}+1,
\end{aligned}
\end{equation}
\begin{equation}
\begin{aligned}\widetilde{b}_{n} & =\left\langle |w_{n}|^{2}\right\rangle +\left\langle s_{n}\right\rangle b_{n}+\left\langle 1-s_{n}\right\rangle \overline{b}_{n}\\
 & =|\mu_{n}|^{2}+\sigma_{n}^{2}+\widetilde{\pi}_{n}b_{n}+\left(1-\widetilde{\pi}_{n}\right)\overline{b}_{n},
\end{aligned}
\end{equation}
 where $\mu_{n}$ and $\sigma_{n}^{2}$ are the posterior mean and
variance of weight $w_{n}$ respectively, and $\widetilde{\pi}_{n}$
is the posterior expectation of $s_{n}$.

\emph{2) Update of $q\left(\mathbf{s}\right)$:}\textbf{ }In subproblem
2, we can achieve the optimal solution $q^{\star}\left(\mathbf{s}\right)$
by letting $q^{\star}\left(\mathbf{s}\right)=\widetilde{p}_{\mathbf{s}}$.
In this case, $q^{\star}\left(\mathbf{s}\right)$ can be derived as
\begin{equation}
q^{\star}\left(\mathbf{s}\right)=\prod_{n=1}^{N}\left(\widetilde{\pi}_{n}\right)^{s_{n}}\left(1-\widetilde{\pi}_{n}\right)^{1-s_{n}},
\end{equation}
 where $\widetilde{\pi}_{n}=\frac{C_{1}}{C_{1}+C_{2}}.$ $C_{1}$
and $C_{2}$ are given by:

\begin{equation}
C_{1}=\frac{\pi_{n}b_{n}^{a_{n}}}{\Gamma\left(a_{n}\right)}e^{\left(a_{n}-1\right)\left\langle \ln\rho_{n}\right\rangle -b_{n}\left\langle \rho_{n}\right\rangle },
\end{equation}
\begin{equation}
C_{2}=\frac{\left(1-\pi_{n}\right)\overline{b}_{n}^{\overline{a}_{n}}}{\Gamma\left(\overline{a}_{n}\right)}e^{\left(\overline{a}_{n}-1\right)\left\langle \ln\rho_{n}\right\rangle -\overline{b}_{n}\left\langle \rho_{n}\right\rangle },
\end{equation}
 where $\left\langle \ln\rho_{n}\right\rangle =\psi\left(\widetilde{a}_{n}\right)-\ln\left(\widetilde{b}_{n}\right)$,
$\psi\left(x\right)=\frac{d}{dx}\ln\left(\Gamma\left(x\right)\right)$
is the digamma function. The detailed derivation of $q^{\star}\left(\boldsymbol{\rho}\right)$
and $q^{\star}\left(\mathbf{s}\right)$ is provided in the Appendix
B.

\emph{3) Update of }\textbf{\emph{$q\left(\mathbf{w}\right)$:}} Expanding
the objective function in subproblem 3, we have 
\begin{equation}
\begin{aligned} & D_{KL}\left(q\left(\mathbf{w}\right)||\widetilde{p}_{\mathbf{w}}\right)\\
 & =\mathbb{E}_{q\left(\mathbf{w}\right)}\ln q\left(\mathbf{w}\right)-\mathbb{E}_{q\left(\mathbf{w}\right)}\left[\left\langle \ln p\left(\mathbf{w},\boldsymbol{\rho},\mathbf{s},\mathcal{D}\right)\right\rangle _{q\left(\boldsymbol{\rho}\right)q\left(\mathbf{s}\right)}\right]\\
 & =\mathbb{E}_{q\left(\mathbf{w}\right)}\left(\ln q\left(\mathbf{w}\right)-\left\langle \ln p\left(\mathbf{w}|\boldsymbol{\rho}\right)\right\rangle _{q\left(\boldsymbol{\rho}\right)}\right)-\mathbb{E}_{q\left(\mathbf{w}\right)}\ln p\left(\mathcal{D}|\mathbf{w}\right)+const.
\end{aligned}
\label{eq:obj-function}
\end{equation}
 Similar to traditional variational Bayesian model compression \cite{Louizos2017,Molchanov2017},
the objective function \eqref{eq:obj-function} contains two parts:
the ``prior part'' $\mathbb{E}_{q\left(\mathbf{w}\right)}\left(\ln q\left(\mathbf{w}\right)-\left\langle \ln p\left(\mathbf{w}|\boldsymbol{\rho}\right)\right\rangle _{q\left(\boldsymbol{\rho}\right)}\right)$
and the ``data part'' $\mathbb{E}_{q\left(\mathbf{w}\right)}\ln p\left(\mathcal{D}|\mathbf{w}\right)$.
The prior part forces the weights to follow the prior distribution
while the data part forces the weights to express the dataset. The
only difference from traditional variational Bayesian model compression
\cite{Louizos2017,Molchanov2017} is that the prior term $\left\langle \ln p\left(\mathbf{w}|\boldsymbol{\rho}\right)\right\rangle _{q\left(\boldsymbol{\rho}\right)}$
is parameterized by $\boldsymbol{\rho}$ and $\boldsymbol{\rho}$
carries the structure captured by the three layer hierarchical prior.
Our aim is to find the optimal $q\left(\mathbf{w}\right)$ that minimizes
$D_{KL}\left(q\left(\mathbf{w}\right)||\widetilde{p}_{\mathbf{w}}\right)$,
subject to the factorized constraint $q\left(\mathbf{w}\right)=\prod_{n=1}^{N}q\left(w_{n}\right)$.
It can be observed that the objective function contains the likelihood
term $\ln p\left(\mathcal{D}|\mathbf{w}\right)$, which can be very
complicated for complex models. Thus, subproblem 3 is non-convex and
it is difficult to obtain the optimal solution. In the following,
we aim at finding a stationary solution $q^{*}\left(\mathbf{w}\right)$
for subproblem 3. There are several challenges in solving subproblem
3, summarized as follows.
\begin{center}
\noindent\fbox{\begin{minipage}[t]{1\columnwidth - 2\fboxsep - 2\fboxrule}%
\textbf{Challenge 1: Closed form for the prior part }$\mathbb{E}_{q\left(\mathbf{w}\right)}\left(\ln q\left(\mathbf{w}\right)-\left\langle \ln p\left(\mathbf{w}|\boldsymbol{\rho}\right)\right\rangle _{q\left(\boldsymbol{\rho}\right)}\right)$\textbf{.}
In traditional variational Bayesian model compression, this expectation
is usually calculated by approximation \cite{Molchanov2017}, which
involves approximating error. To calculate the expectation accurately,
a closed form for this expectation is required.

\textbf{Challenge 2: Low-complexity deterministic approximation for
the data part }$\mathbb{E}_{q\left(\mathbf{w}\right)}\ln p\left(\mathcal{D}|\mathbf{w}\right)$\textbf{.}
In traditional variational Bayesian model compression \cite{Louizos2017,Molchanov2017},
this intractable expectation is usually approximated by Monte Carlo
sampling. However, Monte Carlo approximation has been pointed out
to have high complexity and low robustness. In addition, the variance
of the Monte Carlo gradient estimates is difficult to control \cite{Wu2018}.
Thus, a low-complexity deterministic approximation for the likelihood
term is required.%
\end{minipage}}
\par\end{center}

\textbf{Closed form for $\mathbb{E}_{q\left(\mathbf{w}\right)}\left(\ln q\left(\mathbf{w}\right)-\left\langle \ln p\left(\mathbf{w}|\boldsymbol{\rho}\right)\right\rangle _{q\left(\boldsymbol{\rho}\right)}\right)$.}
To overcome Challenge 1, we propose to use a Gaussian distribution
as the approximate posterior. That is, $q\left(w_{n}\right)=\mathcal{N}\left(\mu_{n},\sigma_{n}^{2}\right)$,
$q\left(\mathbf{w}\right)=\prod_{n=1}^{N}q\left(w_{n}\right)$, $\mu_{n}$
and $\sigma_{n}^{2}$ are the variational parameters that we want
to optimize. Then, since the prior $p\left(\mathbf{w}|\boldsymbol{\rho}\right)$
is also chosen as a Gaussian distribution, $\mathbb{E}_{q\left(\mathbf{w}\right)}\left(\ln q\left(\mathbf{w}\right)-\left\langle \ln p\left(\mathbf{w}|\boldsymbol{\rho}\right)\right\rangle _{q\left(\boldsymbol{\rho}\right)}\right)$
can be written as the KL-divergence between two Gaussian distributions.
Thus, the expectation has a closed form and can be calculated up to
a constant. Specifically, by expanding $p\left(\mathbf{w}|\boldsymbol{\rho}\right)$,
we have 
\begin{equation}
\begin{aligned}\left\langle \ln p\left(\mathbf{w}|\boldsymbol{\rho}\right)\right\rangle  & =\sum_{n=1}^{N}\left\langle \ln\sqrt{\frac{\rho_{n}}{2\pi}}e^{-\frac{w_{n}^{2}}{2}\rho_{n}}\right\rangle \\
 & =\sum_{n=1}^{N}\left\langle \frac{1}{2}\ln\rho_{n}-\frac{w_{n}^{2}}{2}\rho_{n}+const.\right\rangle \\
 & =\sum_{n=1}^{N}\left(\frac{1}{2}\left\langle \ln\rho_{n}\right\rangle -\frac{w_{n}^{2}}{2}\left\langle \rho_{n}\right\rangle +const.\right)\\
 & =\sum_{n=1}^{N}\left(\ln p\left(w_{n}|\left\langle \rho_{n}\right\rangle \right)+const.\right)\\
 & =\ln p\left(\mathbf{w}|\left\langle \boldsymbol{\rho}\right\rangle \right)+const.,
\end{aligned}
\end{equation}
 where we use $\left\langle \cdot\right\rangle $ to replace $\left\langle \cdot\right\rangle _{q\left(\boldsymbol{\rho}\right)}$
or $\left\langle \cdot\right\rangle _{q\left(\rho\right)}$ for simplicity.
Thus, we have 
\begin{multline}
\mathbb{E}_{q\left(\mathbf{w}\right)}\left(\ln q\left(\mathbf{w}\right)-\left\langle \ln p\left(\mathbf{w}|\boldsymbol{\rho}\right)\right\rangle _{q\left(\boldsymbol{\rho}\right)}\right)\\
=\mathbb{E}_{q\left(\mathbf{w}\right)}\left(\ln q\left(\mathbf{w}\right)-\ln p\left(\mathbf{w}|\mathbb{E}\left[\boldsymbol{\rho}\right]\right)\right)+const.\\
=D_{KL}\left(q\left(\mathbf{w}\right)||p\left(\mathbf{w}|\mathbb{E}\left[\boldsymbol{\rho}\right]\right)\right)+const..\label{eq:transform prior part}
\end{multline}
 Since $q\left(\mathbf{w}\right)$ and $p\left(\mathbf{w}|\mathbb{E}\left[\boldsymbol{\rho}\right]\right)$
are both Gaussian distributions, the KL-divergence between them has
a closed form:
\begin{equation}
D_{KL}\left(q\left(\mathbf{w}\right)||p\left(\mathbf{w}|\mathbb{E}\left[\boldsymbol{\rho}\right]\right)\right)=\sum_{n=1}^{N}\left(\ln\frac{\widetilde{\sigma}_{n}}{\sigma_{n}}+\frac{\sigma_{n}^{2}+\mu_{n}^{2}}{2\widetilde{\sigma}_{n}^{2}}-\frac{1}{2}\right),
\end{equation}
 where $\widetilde{\sigma}_{n}^{2}=\frac{1}{\mathbb{E}\left[\rho_{n}\right]}$
is the variance of the prior $p\left(\mathbf{w}|\mathbb{E}\left[\boldsymbol{\rho}\right]\right)$.
Since our aim is to minimize \eqref{eq:obj-function}, the constant
term in \eqref{eq:transform prior part} can be omitted.

\textbf{Low-complexity deterministic approximation for $\mathbb{E}_{q\left(\mathbf{w}\right)}\ln p\left(\mathcal{D}|\mathbf{w}\right)$.
}To overcome Challenge 2, we propose a low-complexity deterministic
approximation for $\mathbb{E}_{q\left(\mathbf{w}\right)}\ln p\left(\mathcal{D}|\mathbf{w}\right)$
based on Taylor expansion.

For simplicity, let $g\left(\mathbf{w}\right)$ denote the complicated
function $\ln p\left(\mathcal{D}|\mathbf{w}\right)$. Specifically,
we want to construct a deterministic approximation for $\mathbb{E}_{\mathbf{w}\sim q\left(\mathbf{w}\right)}g\left(\mathbf{w}\right)$
with the variational parameters $\mu_{n}$ and $\sigma_{n}^{2}$,
$n=1,...,N$. Following the reparameterization trick in \cite{Molchanov2017}
and \cite{Kingma2015a}, we represent $w_{n}$ by $\mu_{n}+\xi_{n}\sigma_{n}^{2}$,
$\xi_{n}\sim\mathcal{N}\left(0,1\right)$. In this way, $w_{n}$ is
transformed into a deterministic differentiable function of a non-parametric
noise $\xi_{n}$, and $\mathbb{E}_{\mathbf{w}\sim q\left(\mathbf{w}\right)}g\left(\mathbf{w}\right)$
is transformed into $\mathbb{E}_{\boldsymbol{\xi}\sim\mathcal{N}\left(0,1\right)}g\left(\boldsymbol{\mu}+\boldsymbol{\xi\sigma^{2}}\right)$.
Note that in traditional variational Bayesian model compression, this
expectation is approximated by sampling $\boldsymbol{\xi}$. However,
to construct a deterministic approximation, we take the first-order
Taylor approximation of $\mathbb{E}_{\boldsymbol{\xi}\sim\mathcal{N}\left(0,1\right)}g\left(\boldsymbol{\mu}+\boldsymbol{\xi\sigma^{2}}\right)$
at the point $\boldsymbol{\xi}=\mathbb{E}\left[\boldsymbol{\xi}\right]=\mathbf{0}$,
and we have 
\begin{equation}
\mathbb{E}_{\boldsymbol{\xi}}g\left(\boldsymbol{\mu}+\boldsymbol{\xi\sigma^{2}}\right)\approx g\left(\boldsymbol{\mu}+\mathbb{E}\left[\boldsymbol{\xi}\right]\cdot\boldsymbol{\sigma^{2}}\right)=g\left(\boldsymbol{\mu}\right).
\end{equation}
 That is, we use the first-order Taylor approximation to replace $\mathbb{E}_{\boldsymbol{\xi}}g\left(\boldsymbol{\mu}+\boldsymbol{\xi\sigma^{2}}\right)$.
Simulation results show that our proposed approximation can achieve
good performance. Additionally, since our proposed approximation is
deterministic, the gradient variance can be eliminated.

Based on the aforementioned discussion , subproblem 3 can be solved
by solving an approximated problem as follows.

\textbf{Approx. Subproblem 3:} 
\begin{equation}
\begin{aligned}\min_{\boldsymbol{\mu},\boldsymbol{\sigma}^{2}} & \sum_{n=1}^{N}\left(\ln\frac{\widetilde{\sigma}_{n}}{\sigma_{n}}+\frac{\sigma_{n}^{2}+\mu_{n}^{2}}{2\widetilde{\sigma}_{n}^{2}}-\frac{1}{2}\right)-\ln p\left(\mathcal{D}|\boldsymbol{\mu}\right).\end{aligned}
\label{eq:approx subproblem 3}
\end{equation}
 This problem is equivalent to training a Bayesian neural network
with \eqref{eq:approx subproblem 3} as the loss function.

\emph{4) Convergence of Sparse VBI:} The sparse VBI in Module A is
actually a BCD algorithm to solve the problem in \eqref{eq:min_neg_ELBO}.
By the physical meaning of ELBO and KL divergence, the objective function
is continuous on a compact level set. In addition, the objective function
has a unique minimum for two coordinate blocks (\textbf{$q\left(\boldsymbol{\rho}\right)$}
and \textbf{$q\left(\mathbf{s}\right)$}) \cite{Liu2020,Tzikas2008}.
Since we can achieve a stationary point for the third coordinate block
$q\left(\mathbf{w}\right)$, the objective function is regular at
the cluster points generated by the BCD algorithm. Based on the above
discussion, our proposed BCD algorithm satisfies the BCD algorithm
convergence requirements in \cite{Tseng2001}. Thus, we have the following
convergence lemma.
\begin{lem}
(Convergence of Sparse VBI): Every cluster point $q^{*}\left(\mathbf{w},\boldsymbol{\rho},\mathbf{s}\right)=q^{\star}\left(\boldsymbol{\rho}\right)q^{\star}\left(\mathbf{s}\right)q^{*}\left(\mathbf{w}\right)$
generated by the sparse VBI algorithm is a stationary solution of
problem \eqref{eq:min_neg_ELBO}.
\end{lem}

\subsection{Message Passing (Module B)}

In Module B, SPMP is performed on the support factor graph $\mathcal{G}_{s}$
and the normalized message $v_{s_{n}\rightarrow h_{B}}$ is fed back
to Module A as the prior probability $\hat{p}\left(s_{n}\right)$.

\begin{figure}[tb]
\begin{centering}
\includegraphics[width=0.4\textwidth]{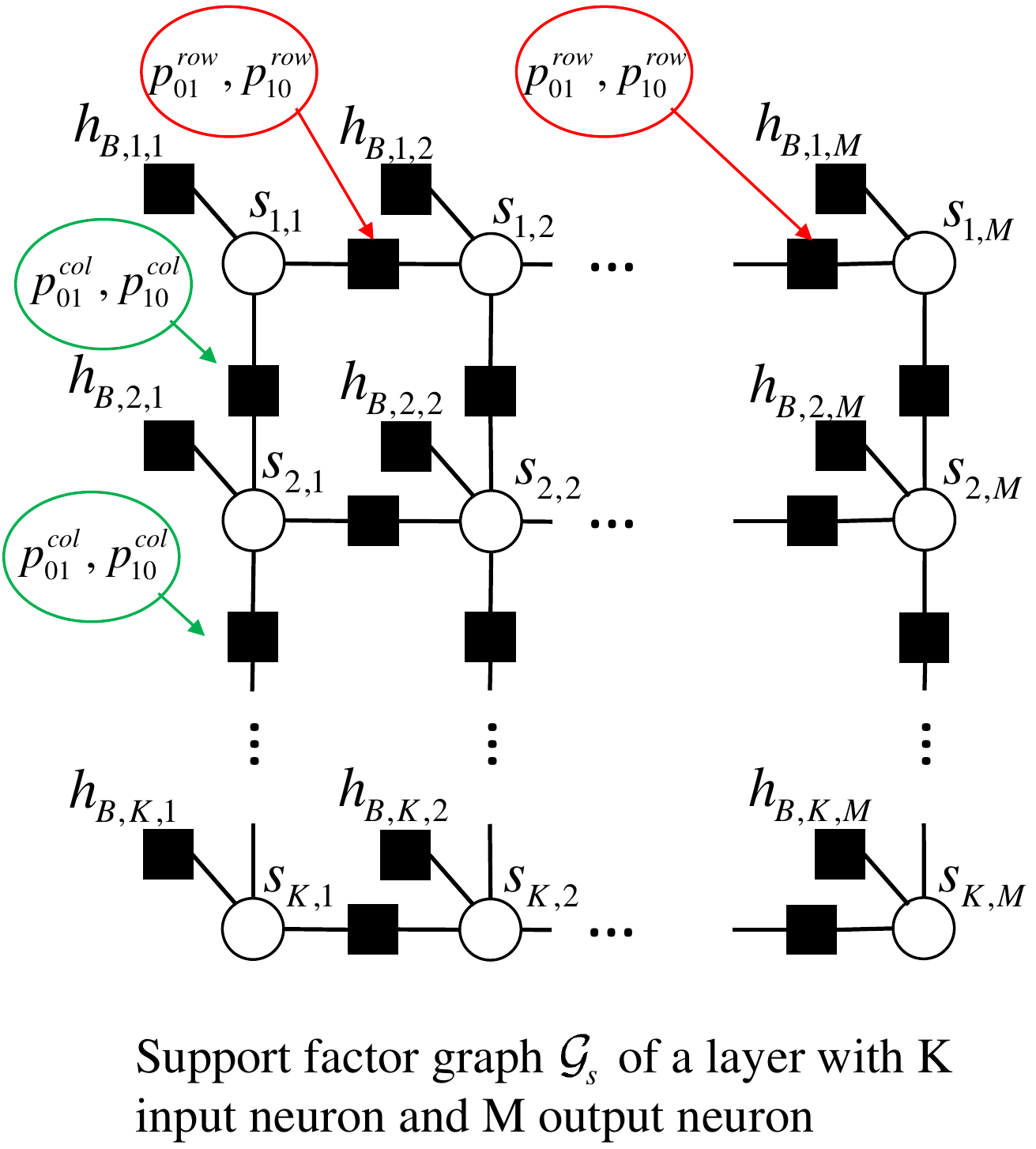}
\par\end{centering}
\caption{Illustration of the support factor graph in Module B with $K$ input
neurons and $M$ output neurons.\label{fig:Illustration of the Module B}}
\end{figure}

\begin{algorithm}[tb]
\begin{algorithmic}[1] 
\Require{training set $\mathcal{D}$, priors $p\left(\mathbf{w}\right)$, $p\left(\boldsymbol{\rho}\right)$, $p\left(\mathbf{s}\right)$, maximum iteration number $I_{max}$.} 
\Ensure{$\mathbf{w}^{*}$, $\boldsymbol{\rho}^{*}$, $\mathbf{s}^{*}$.}
\State{Initialize $\pi_{n}$.}
\For{$k=1,...,I_{m}$} 
\State{\textbf{Module A:}}
\State{Initialize the variational distribution $q\left(\boldsymbol{\rho}\right)$ and $q\left(\mathbf{s}\right)$.}
\While{not converged}
\State{Update $q\left(\boldsymbol{\rho}\right)$, $q\left(\mathbf{s}\right)$ and $q\left(\mathbf{w}\right)$.}
\EndWhile
\State{Calculate $v_{\eta_{n}\rightarrow s_{n}}$ based on $(17)$ and send $v_{\eta_{n}\rightarrow s_{n}}$ to Module B.}
\State{\textbf{Module B:}}
\State{Perform SPMP on $\mathcal{G}_{s}$, send $v_{h\rightarrow s_{n}}$ to   Module A.}
\If{converge}
\State{\textbf{break}}
\EndIf
\State{$k=k+1$.}
\EndFor
\State{Output $\mathbf{w}^{*}=\arg\max_{\mathbf{w}}q^{*}\left(\mathbf{w}\right)$, $\boldsymbol{\rho}^{*}=\arg\max_{\boldsymbol{\rho}}q^{\star}\left(\boldsymbol{\rho}\right)$, $\mathbf{s}^{*}=\arg\max_{\mathbf{s}}q^{\star}\left(\mathbf{s}\right)$.}
\end{algorithmic}

\caption{Turbo-VBI Algorithm for Model Compression}
\end{algorithm}

We focus on one fully connected layer with $K$ input neurons and
$M$ output neurons for easy elaboration. The support prior $p\left(\mathbf{s}\right)$
for this $K\times M$ layer is modeled as an MRF. The factor graph
$\mathcal{G}_{s}$ is illustrated in Fig. \ref{fig:Illustration of the Module B}.
Here we use $s_{i,j}$ to denote the support variable in the $i$-th
row and $j$-th column. The unary factor node $h_{B,i,j}$ carries
the probability given by \eqref{eq:hbn}. The pair-wise factor nodes
carry the row and column transition probability $p\left(s_{row,m+1}|s_{row,m}\right)$
and $p\left(s_{col,k+1}|s_{col,k}\right)$, respectively. The parameters
$p_{01}^{row},p_{10}^{row}$ and $p_{01}^{col},p_{10}^{col}$ for
the MRF model are given in Section \ref{subsec:Three-Layer-Hierarchical-Prior}.
Let $v_{s\rightarrow f}\left(s\right)$ denote the message from the
variable node $s$ to the factor node $f$, and $v_{f\rightarrow s}\left(s\right)$
denote the message from the factor node $f$ to the variable node
$s$. According to the sum-product law, the messages are updated as
follows:
\begin{equation}
v_{s\rightarrow f}\left(s\right)=\prod_{h\in n\left(s\right)\setminus\{f\}}v_{h\rightarrow s}\left(s\right),
\end{equation}
\begin{equation}
v_{f\rightarrow s}\left(s\right)=\begin{cases}
f\left(s\right) & f\;is\;unary\;factor\;node,\\
f\left(s\right)\cdot v_{t\rightarrow f}\left(t\right) & f\;is\;pairwise\;factor\;node,
\end{cases}
\end{equation}
where $n\left(s\right)\setminus\{f\}$ denotes the neighbour factor
nodes of $s$ except node $f$, and $t$ denotes the other neighbour
variable node of $f$ except node $s$. After the SPMP algorithm is
performed, the final message $v_{s\rightarrow h_{B}}$ from each variable
node $s_{i,j}$ to the connected unary factor node $h_{B,i,j}$ is
sent back to Module A.

The overall algorithm is summarized in Algorithm 1.

\emph{Remark 2. (Comparison with traditional variational inference)}
Traditional variational inference cannot handle the proposed three-layer
sparse prior $p\left(\mathbf{w},\boldsymbol{\rho},\mathbf{s}\right)$
while our proposed Turbo-VBI is specially designed to handle it. In
existing Bayesian compression works \cite{van2020bayesian}\nocite{Molchanov2017}-\cite{Louizos2017},
the KL-divergence between the true posterior $p\left(\mathbf{w}|\mathcal{D}\right)$
and the variational posterior $q\left(\mathbf{w}\right)$ is easy
to calculate, so that the KL-divergence can be directly optimized.
Thus, simple variational inference can be directly applied. The fundamental
reason behind this is that the prior $p\left(\mathbf{w}\right)$ in
these works are relatively simple, and they usually only have one
layer \cite{van2020bayesian,Molchanov2017} or two layers \cite{Louizos2017}.
Thus, the KL-divergence between the prior $p\left(\mathbf{w}\right)$
and the variational posterior $q\left(\mathbf{w}\right)$ is easy
to calculate, which is a must in calculating the KL-divergence between
the true posterior $p\left(\mathbf{w}|\mathcal{D}\right)$ and the
variational posterior $q\left(\mathbf{w}\right)$. However, these
simple priors cannot promote the desired regular structure in the
weight matrix and lack the flexibility to configure the pruned structure.
Thus, we introduce the extra support layer $\mathbf{s}$, which leads
to the three layer hierarchical prior $p\left(\mathbf{w},\boldsymbol{\rho},\mathbf{s}\right)$.
With this three-layer prior, traditional variational inference cannot
be applied because the KL-divergence between the prior $p\left(\mathbf{w},\boldsymbol{\rho},\mathbf{s}\right)$
and the variational posterior $q\left(\mathbf{w},\boldsymbol{\rho},\mathbf{s}\right)$
is difficult to calculate. In order to inference the posterior, we
propose to separate the factor graph into two parts and iteratively
perform Bayesian inference, which leads to our proposed Turbo-VBI
algorithm.

\section{Performance Analysis\label{sec:Performance-Analysis}}

In this section, we evaluate the performance of our proposed Turbo-VBI
based model compression method on some standard neural network models
and datasets. We also compare with several baselines, including classic
and state-of-the-art model compression methods. The baselines considered
are listed as follows:
\begin{enumerate}
\item \emph{Variational dropout (VD):} This is a very classic method in
the area of Bayesian model compression, which is proposed in \cite{Molchanov2017}.
A single-layer improper log-uniform prior is assigned to each weight
to induce sparsity during training.
\item \emph{Group variational dropout (GVD):} This is proposed in \cite{Louizos2017},
which can be regarded as an extension of VD into group pruning. In
this algorithm, each weight follows a zero-mean Gaussian prior. The
prior scales of the weights in the same group jointly follow an improper
log-uniform distribution or a proper half-Cauchy distribution to realize
group pruning. In this paper, we choose the improper log-uniform prior
for comparison. Note that some more recent Bayesian compression methods
mainly focus on a quantization perspective \cite{van2020bayesian,yuan2019enhanced},
which is a different scope from our work. Although our work can be
further extended into quantization area, we only focus on structured
pruning here. To our best knowledge, GVD is still one of the state-of-the-art
Bayesian pruning methods without considering quantization.
\item \emph{Polarization regularizer based pruning (polarization):} This
is proposed in \cite{zhuang2020neuron}. The algorithm can push some
groups of weights to zero while others to larger values by assigning
a polarization regularizer to the batch norm scaling factor of each
weight group
\[
\begin{aligned}\min_{\mathbf{w}} & \frac{1}{D}\sum_{d=1}^{D}L\left(NN\left(x_{d},\mathbf{w}\right),y_{d}\right)+R\left(\mathbf{w}\right)\\
 & +\lambda\left(t\left\Vert \boldsymbol{\gamma}\right\Vert _{1}-\left\Vert \boldsymbol{\gamma}-\bar{\gamma}\mathbf{1}_{n}\right\Vert _{1}\right),
\end{aligned}
\]
 where $\lambda\left(t\left\Vert \boldsymbol{\gamma}\right\Vert _{1}-\left\Vert \boldsymbol{\gamma}-\bar{\gamma}\mathbf{1}_{n}\right\Vert _{1}\right)$
is the proposed polarization regularizer and $\boldsymbol{\gamma}$
is the batch norm scaling factor of each weight.
\item \emph{Single-shot network pruning (SNIP):} This is proposed in \cite{Lee2018}.
The algorithm measures the weight importance by introducing an auxiliary
variable to each weight. The importance of each weight is achieved
before training by calculating the gradient of its ``importance variable''.
After this, the weights with less importance are pruned before the
normal training.
\end{enumerate}
We perform experiments on LeNet-5, AlexNet, and VGG-11 architectures,
and use the following benchmark datasets for the performance comparison. 
\begin{enumerate}
\item Fashion-MNIST \cite{xiao2017fashion}: It consists of a training set
of 60000 examples and a test set of 10000 examples. Each example is
a $28\times28$ grayscale image of real life fashion apparel, associated
with a label from 10 classes.
\item CIFAR-10 \cite{KrizhevskyaccessedAugust212022}: It is the most widely
used dataset to compare neural network pruning methods. It consists
of a training set of 50000 examples and a test set of 10000 examples.
Each example is a three-channel $32\times32$ RGB image of real life
object, associated with a label from 10 classes.
\item CIFAR-100 \cite{KrizhevskyaccessedAugust212022}: It is considered
as a harder version of CIFAR-10 dataset. Instead of 10 classes in
CIFAR-10, it consists of 100 classes with each class containing 600
images. There are 500 training images and 100 testing images in each
class.
\end{enumerate}
We will compare our proposed Turbo-VBI based model compression method
with the baselines from different dimensions, including 1) an overall
performance comparison; 2) visualization of the weight matrix structure;
3) CPU and GPU acceleration and 4) robustness of the pruned neural
network.

We run LeNet-5 on Fashion-MNIST, and AlexNet on CIFAR-10 and VGG on
CIFAR-100. Throughout the experiments, we set the learning rate as
0.01, and $a=b=\overline{a}=1,\overline{b}=1\times10^{-3}$. For LeNet-5+Fashion-MNIST,
we set the batch size as 64, and for AlexNet+CIFAR-10 and VGG+CIFAR-100,
we set the batchsize as 128 . All the methods except for SNIP go through
a fine tuning after pruning. For LeNet, the fine tuning is 15 epochs
while for AlexNet and VGG, the fine tuning is 30 epochs. Note that
the models and parameter settings may not achieve state-of-the-art
accuracy but it is enough for comparison.

\subsection{Overall Performance Comparison\label{subsec:Overall-Performance-Comparison}}
\begin{center}
\begin{table*}[tb]
\subfloat[Results on LeNet+Fashion-MNIST.\label{tab:Performance-comparison-lenet}]{\begin{centering}
\begin{tabular}{>{\centering}p{0.15\textwidth}>{\centering}p{0.15\textwidth}>{\centering}p{0.15\textwidth}>{\centering}p{0.15\textwidth}>{\centering}p{0.15\textwidth}>{\centering}p{0.15\textwidth}}
\hline 
Network & Method & Accuracy (\%) & $\frac{|\mathbf{w}\neq0|}{|\mathbf{w}|}$ (\%) & Pruned structure & FLOPs reduction (\%)\tabularnewline
\hline 
 & No pruning & 89.01 & 100 & 6-16-120-84 & 0\tabularnewline
\cline{2-6} \cline{3-6} \cline{4-6} \cline{5-6} \cline{6-6} 
 & VD & \textbf{88.91} & 6.16 & 6-15-49-51 & 11.82\tabularnewline
\cline{2-6} \cline{3-6} \cline{4-6} \cline{5-6} \cline{6-6} 
LeNet-5 & GVD & 88.13 & 4.20 & 5-7-21-23 & 53.89\tabularnewline
\cline{2-6} \cline{3-6} \cline{4-6} \cline{5-6} \cline{6-6} 
6-16-120-84 & Polarization & 87.27 & 17.96 & 4-8-46-27 & 59.04\tabularnewline
\cline{2-6} \cline{3-6} \cline{4-6} \cline{5-6} \cline{6-6} 
 & SNIP & 82.83 & 9.68 & 5-16-73-57 & 19.82\tabularnewline
\cline{2-6} \cline{3-6} \cline{4-6} \cline{5-6} \cline{6-6} 
 & Proposed & 88.77 & \textbf{1.14} & \textbf{3-5-16-17} & \textbf{76.03}\tabularnewline
\cline{2-6} \cline{3-6} \cline{4-6} \cline{5-6} \cline{6-6} 
\end{tabular}
\par\end{centering}
}\hfill{}\subfloat[Results on AlexNet+CIFAR-10.\label{tab:Results-on-AlexNet+CIFAR-10.}]{%
\begin{tabular}{>{\centering}p{0.15\textwidth}>{\centering}p{0.15\textwidth}>{\centering}p{0.15\textwidth}>{\centering}p{0.15\textwidth}>{\centering}p{0.15\textwidth}>{\centering}p{0.15\textwidth}}
\hline 
Network & Method & Accuracy (\%) & $\frac{|\mathbf{w}\neq0|}{|\mathbf{w}|}$ (\%) & Pruned structure & FLOPs reduction (\%)\tabularnewline
\hline 
 & No pruning & 66.18 & 100 & 6-16-32-64-128-120-84 & 0\tabularnewline
\cline{2-6} \cline{3-6} \cline{4-6} \cline{5-6} \cline{6-6} 
 & VD & 64.91 & \textbf{3.17} & 6-16-31-61-117-42-74 & 6.12\tabularnewline
\cline{2-6} \cline{3-6} \cline{4-6} \cline{5-6} \cline{6-6} 
AlexNet & GVD & 64.82 & 9.30 & 6-16-27-31-21-17-14 & 39.29\tabularnewline
\cline{2-6} \cline{3-6} \cline{4-6} \cline{5-6} \cline{6-6} 
6-16-32-64-128-120-84 & Polarization & 63.95 & 25.95 & 4-15-16-20-19-58-53 & \textbf{65.27}\tabularnewline
\cline{2-6} \cline{3-6} \cline{4-6} \cline{5-6} \cline{6-6} 
 & SNIP & 63.19 & 11.75 & 6-16-30-57-102-56-68 & 12.51\tabularnewline
\cline{2-6} \cline{3-6} \cline{4-6} \cline{5-6} \cline{6-6} 
 & Proposed & \textbf{65.57} & 3.42 & \textbf{6-16-25-20-7-7-7} & 45.94\tabularnewline
\cline{2-6} \cline{3-6} \cline{4-6} \cline{5-6} \cline{6-6} 
\end{tabular}}\hfill{}\subfloat[Results on VGG11+CIFAR-100.\label{tab:Results-on-VGG11+CIFAR-100.}]{%
\begin{tabular}{>{\centering}p{0.15\textwidth}>{\centering}p{0.15\textwidth}>{\centering}p{0.15\textwidth}>{\centering}p{0.15\textwidth}>{\centering}p{0.15\textwidth}>{\centering}p{0.15\textwidth}}
\hline 
Network & Method & Accuracy (\%) & $\frac{|\mathbf{w}\neq0|}{|\mathbf{w}|}$ (\%) & Pruned structure & FLOPs reduction (\%)\tabularnewline
\hline 
 & No pruning & 68.28 & 100 & 64-128-256-256-512-512-512-512-100 & 0\tabularnewline
\cline{2-6} \cline{3-6} \cline{4-6} \cline{5-6} \cline{6-6} 
 & VD & \textbf{67.24} & 5.39 & 64-128-256-211-447-411-293-279-44 & 16.63\tabularnewline
\cline{2-6} \cline{3-6} \cline{4-6} \cline{5-6} \cline{6-6} 
VGG-11 & GVD & 66.52 & 9.05 & 63-112-204-134-271-174-91-77-33 & 60.90\tabularnewline
\cline{2-6} \cline{3-6} \cline{4-6} \cline{5-6} \cline{6-6} 
64-128-256-256-512-512-512-512-100 & Polarization & 64.41 & 26.74 & 62-122-195-148- 291-133-76-74-68 & 59.54\tabularnewline
\cline{2-6} \cline{3-6} \cline{4-6} \cline{5-6} \cline{6-6} 
 & SNIP & 65.20 & 17.79 & 64-128-254-246-441-479-371-274-69 & 7.92\tabularnewline
\cline{2-6} \cline{3-6} \cline{4-6} \cline{5-6} \cline{6-6} 
 & Proposed & 67.12 & \textbf{4.68} & \textbf{60-119-191-148- 181-176-44-64-59} & \textbf{63.14}\tabularnewline
\cline{2-6} \cline{3-6} \cline{4-6} \cline{5-6} \cline{6-6} 
\end{tabular}

}\caption{Overall performance comparisons to the baselines on different network
models and datasets. Top-1 accuracy is reported. The best results
are bolded.\label{tab:Overall-performance-comparisons}}
\end{table*}
\par\end{center}

We first give an overall performance comparison of different methods
in terms of accuracy, sparsity rate, pruned structure and FLOPs reduction.
The results are summarized in Table \ref{tab:Overall-performance-comparisons}.
The structure is measured by output channels for convolutional layers
and neurons for fully connected layers. It can be observed that our
proposed method can achieve an extremely low sparsity rate, which
is at the same level (or even lower than) the unstructured pruning
methods (VD and SNIP). Moreover, it can also achieve an even more
compact pruned structure than the structured pruning methods (polarization
and GVD). At the same time, our proposed method can maintain a competitive
accuracy. We summarize that the superior performance of our proposed
method comes from two aspects: 1) superior individual weight pruning
ability, which leads to low sparsity rate, and 2) superior neuron
pruning ability, which leads to low FLOPs. Compared to random pruning
methods like VD and SNIP which generally can achieve lower sparsity
rate than group pruning methods, our proposed method can achieve even
lower sparsity rate. This shows that our proposed sparse prior has
high efficiency in pruning individual weights. This is because the
regularization ability of $p\left(\mathbf{w|\boldsymbol{\rho}}\right)$
can be configured by setting small $\overline{b}$. Compared to group
pruning methods like GVD and polarization, our proposed method can
achieve a more compact structure. This shows our proposed method is
also highly efficient in pruning neurons. The reason is two fold:
First, each row in the prior $p\left(\mathbf{s}\right)$ can be viewed
as a Markov chain. When some weights of a neuron are pruned, the other
weights will also have a very high probability to be pruned. Second,
the surviving weights in a weight matrix tend to gather in clusters,
which means the surviving weights of different neurons in one layer
tend to activate the same neurons in the next layer. This greatly
improves the regularity from the input side, while the existing group
pruning methods like GVD and polarization can only regularize the
output side of a neuron.Visualization of Pruned Structure

\begin{figure}[tb]
\begin{centering}
\includegraphics[width=0.45\textwidth]{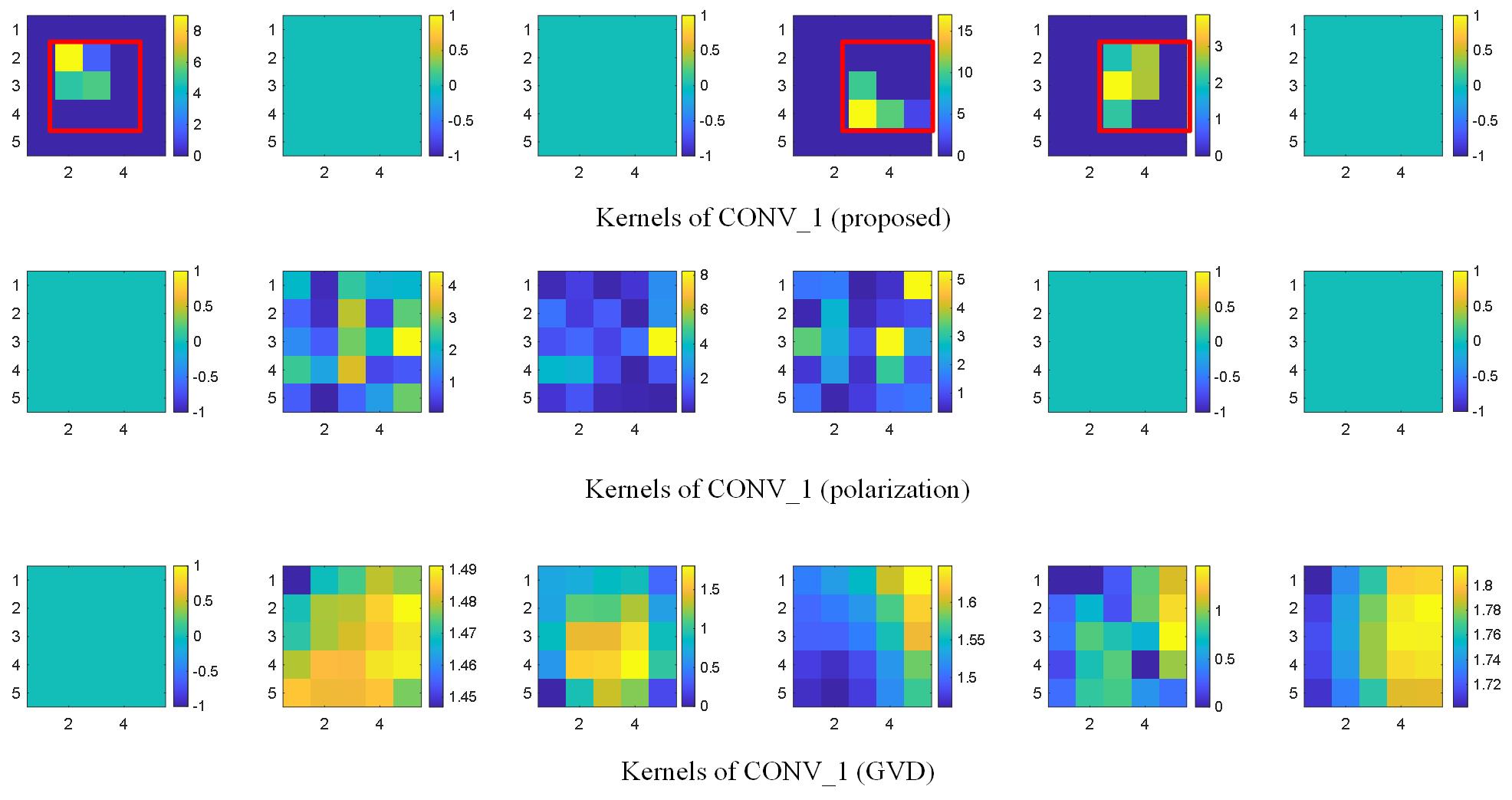}
\par\end{centering}
\caption{Kernel structures of CONV\_1 in LeNet.\label{fig:Kernel-structures}}
\end{figure}

In this subsection, we visualize the pruned weight matrix of our proposed
method and compare to the baselines to further elaborate the benefits
brought by the proposed regular structure. Fig. \ref{fig:Kernel-structures}
illustrates the 6 kernels in the first layer of LeNet. We choose GVD
and polarization for comparison here because these two are also structured
pruning methods. It can be observed that GVD and polarization can
both prune the entire kernels but the weights in the remaining kernels
cannot be pruned and have no structure at all. However, in our proposed
method, the weights in the remaining kernels are clearly pruned in
a structured manner. In particular, we find that the three remaining
$5\times5$ kernels all fit in smaller kernels with size $3\times3$.
This clearly illustrates the clustered structure promoted by the MRF
prior. With the clustered pruning, we can directly replace the original
large kernels with smaller kernels, and enjoy the storage and computation
benefits of the smaller kernel size.

\subsection{Acceleration Performance}

\begin{figure}[tb]
\begin{centering}
\subfloat[Avg. inferencing time on CPU.]{\centering{}\includegraphics[width=0.45\textwidth]{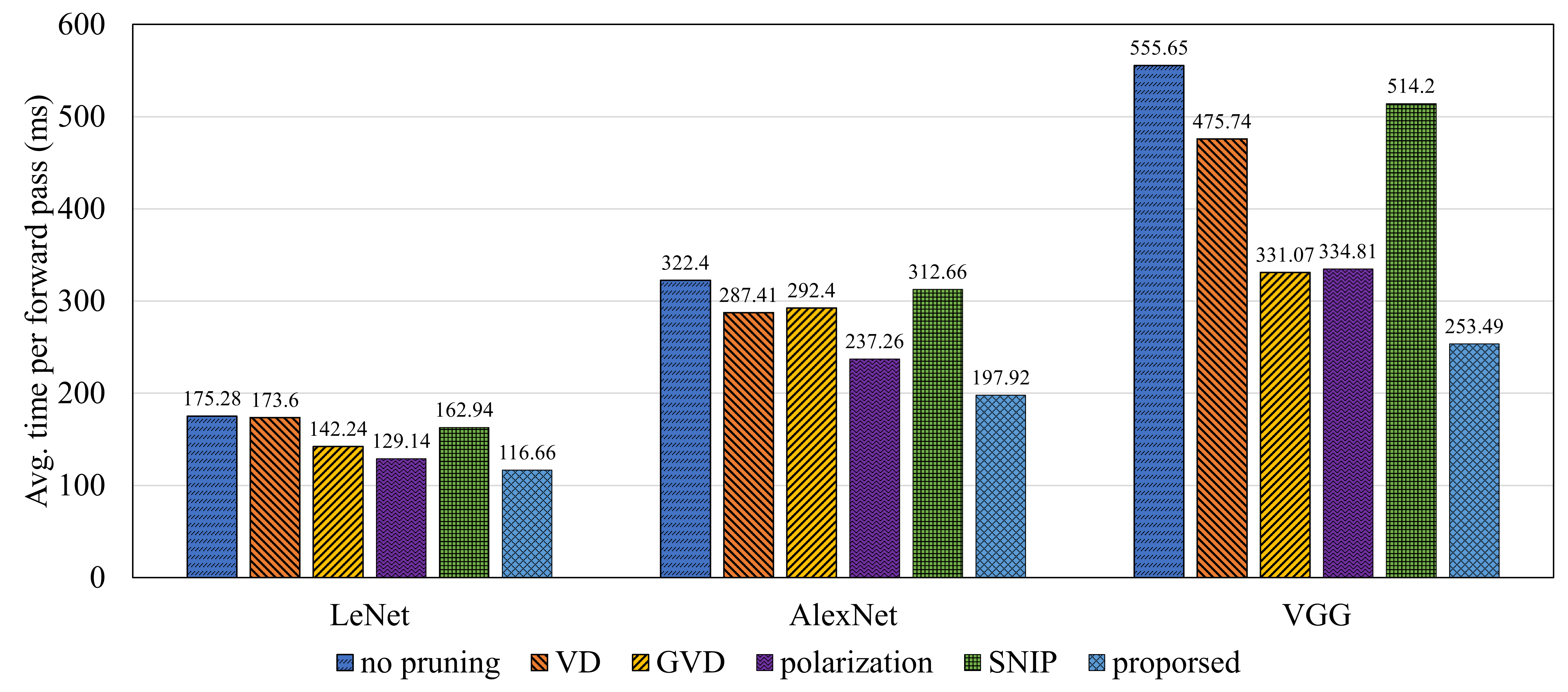}}\hfill{}\subfloat[Avg. inferencing time on GPU.]{\begin{centering}
\includegraphics[width=0.45\textwidth]{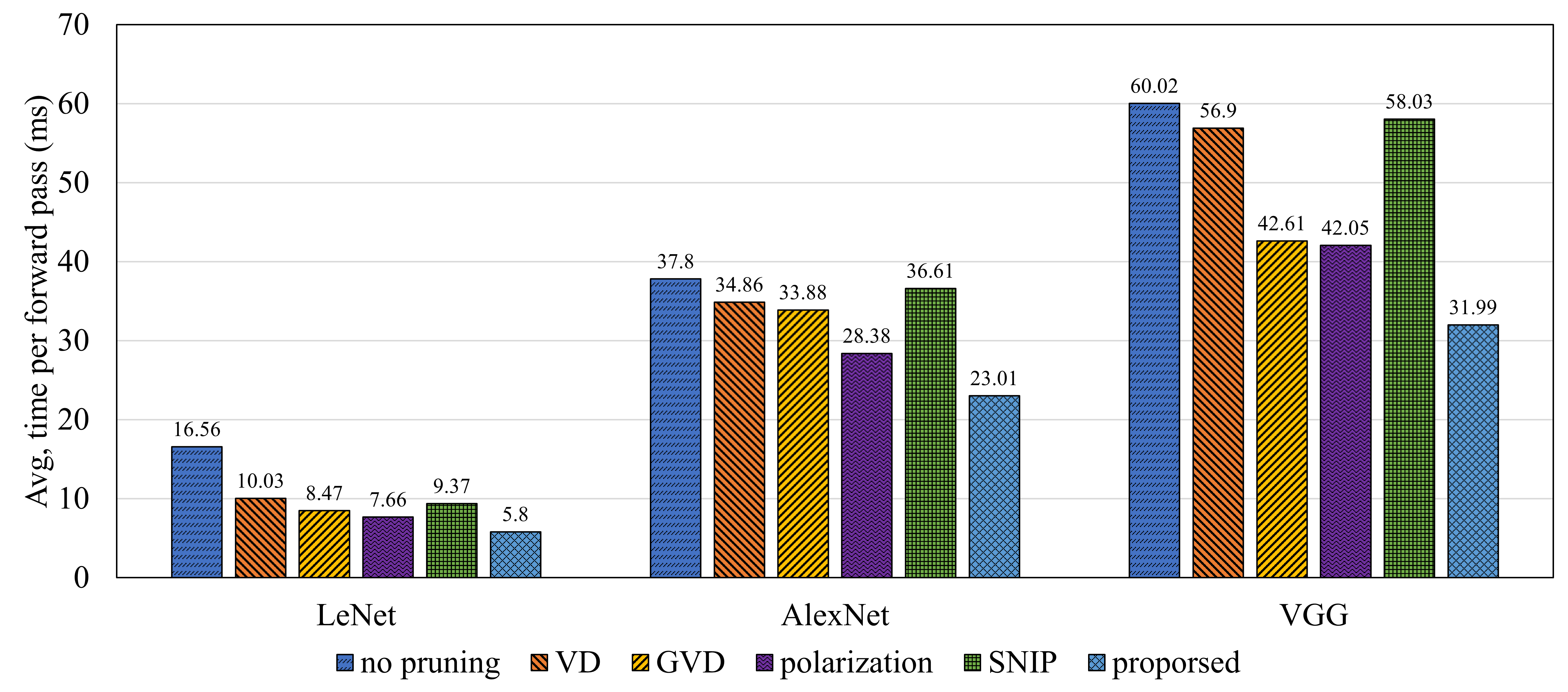}
\par\end{centering}
}
\par\end{centering}
\caption{Avg. time costed by one forward pass of a batch on CPU and GPU. For
LeNet and AlexNet, the batch size is 10000 examples and each results
is averaged on 1000 experiments. For VGG, the batch size is 1000 examples
and each result is averaged on 100 experiments.\label{fig:Avg.-time-costed}}
\end{figure}

In this subsection, we compare the acceleration performance of our
proposed method to the baselines. Note that the primary goal of the
structured pruning is to reduce the computational complexity and accelerate
the inference of the neural networks. In the weight matrix of our
proposed method, we record the clusters larger than $3\times3$ as
an entirety. That is, we record the location and size of the clusters
instead of recording the exact location for each element. In Fig.
\ref{fig:Avg.-time-costed}, we plot the average time costed by one
forward pass of a batch on CPU and GPU respectively. It can be observed
that our proposed method can achieve the best acceleration performance
on different models. For LeNet, our proposed method can achieve a
$1.50\times$ gain on CPU and $2.85\times$ gain on GPU. For AlexNet,
the results are $1.63\times$ and $1.64\times$. For VGG, the results
are $2.19\times$ and $1.88\times$. Note that the unstructured pruning
methods show little acceleration gain because their sparse structure
is totally random. We summarize the reason for our acceleration gain
from two aspects: 1) More efficient neuron pruning. As shown in subsection
\ref{subsec:Overall-Performance-Comparison}, our proposed method
can result in a more compact model and thus leading to a larger FLOPs
reduction. 2) More efficient decoding of weight matrix. Existing methods
require to store the exact location for each unpruned weight, no matter
in CSR, CSC or COO format because the sparse structure is irregular.
However, in our proposed method, we only need the location and size
for each cluster to decode all the unpruned weights in the cluster.
Thus, the decoding is much faster. Note that the performance of our
proposed method can be further improved if the structure is explored
during the matrix-matrix multiplication. For example, the zero blocks
can be skipped in matrix tiling, as discussed in subsection \ref{subsec:Overall-Performance-Comparison}.
However, this involves modification in lower-level CUDA codes so we
do not study it here.

\subsection{Insight Into the Proposed Algorithm}

\begin{figure}[tb]
\begin{centering}
\includegraphics[width=0.45\textwidth]{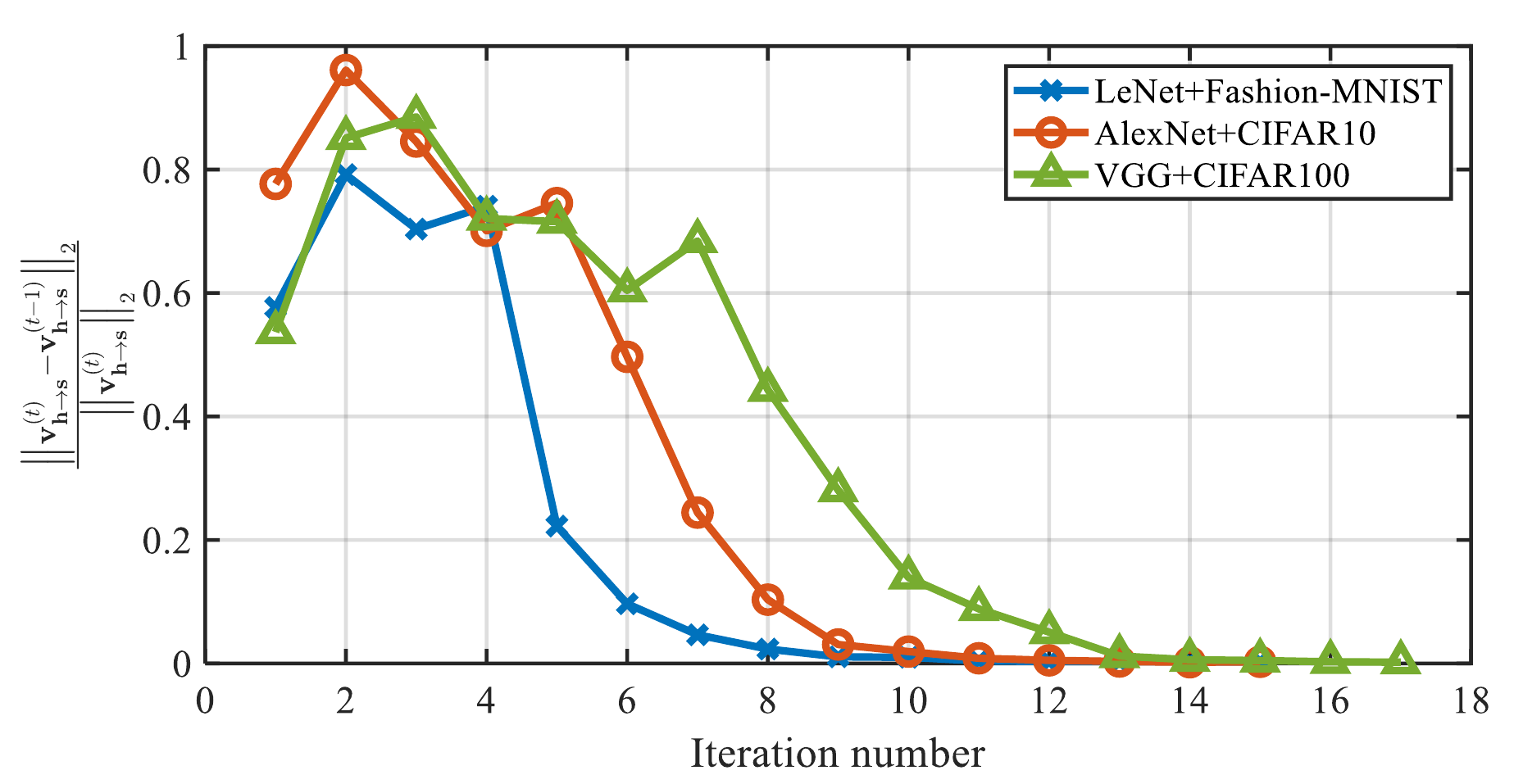}
\par\end{centering}
\caption{Convergence of proposed algorithm. The convergence if measured by
the change in message $\boldsymbol{v}_{h\rightarrow s_{n}}$.\label{fig:Convergence-of-proposed}}
\end{figure}

\begin{figure*}[tb]
\begin{centering}
\includegraphics[width=0.9\textwidth]{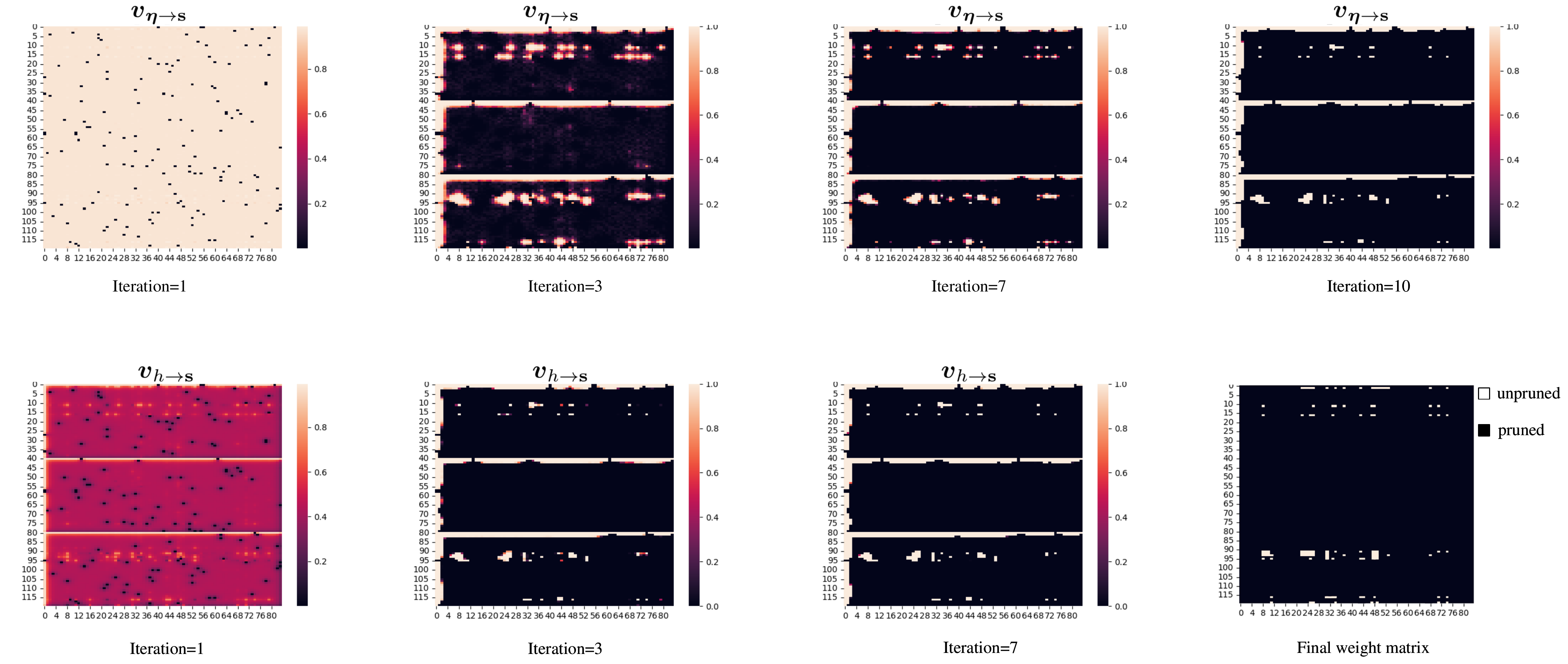}
\par\end{centering}
\caption{Visualization of the message $\boldsymbol{v}_{h\rightarrow s_{n}}$
and $\boldsymbol{v}_{\eta_{n}\rightarrow s_{n}}$ for the Dense\_3
layer of AlexNet. $\boldsymbol{v}_{h\rightarrow s_{n}}$ carries the
prior information of support from Module B and $\boldsymbol{v}_{\eta_{n}\rightarrow s_{n}}$
carries the posterior information of support from Module A. It can
be clearly observed that the structured prior captured from Module
B gradually regularizes the posterior. Eventually both $\boldsymbol{v}_{h\rightarrow s_{n}}$
and $\boldsymbol{v}_{\eta_{n}\rightarrow s_{n}}$ converge to a stable
state.\label{fig:Visualization of message}}
\end{figure*}

In this subection, we provide some insight into the proposed method.
First, Fig. \ref{fig:Convergence-of-proposed} illustrates the convergence
behavior of our proposed method on different tasks. It can be observed
that generally our proposed method can converge in 15 iterations in
all the tasks. Second, we illustrate how the structure is gradually
captured by the prior information $\boldsymbol{v}_{h\rightarrow s_{n}}$
from Module B and how the posterior in Module A is regularized to
this structure. As the clustered structure in kernels has been illustrated
in Fig. \ref{fig:Kernel-structures}, here in Fig. \ref{fig:Visualization of message},
we visualize the message $\boldsymbol{v}_{h\rightarrow s_{n}}$ and
$\boldsymbol{v}_{\eta_{n}\rightarrow s_{n}}$ of the FC\_3 layer in
AlexNet. It can be observed that in iteration 1, the message $\boldsymbol{v}_{\eta_{n}\rightarrow s_{n}}$
has no structure because the prior $\hat{p}\left(\mathbf{s}\right)$
in Module A is randomly initialized. During the iterations of the
algorithm, $\boldsymbol{v}_{\eta_{n}\rightarrow s_{n}}$ from Module
B gradually captures the structure from the MRF prior and acts as
a regularization in Module A, so that structure is gradually promoted
in $\boldsymbol{v}_{h\rightarrow s_{n}}$. Finally, the support matrix
has a clustered structure and the weight matrix also has a similar
structure.

\subsection{Robustness of Pruned Model}

\begin{figure}[tb]
\begin{centering}
\includegraphics[width=0.45\textwidth]{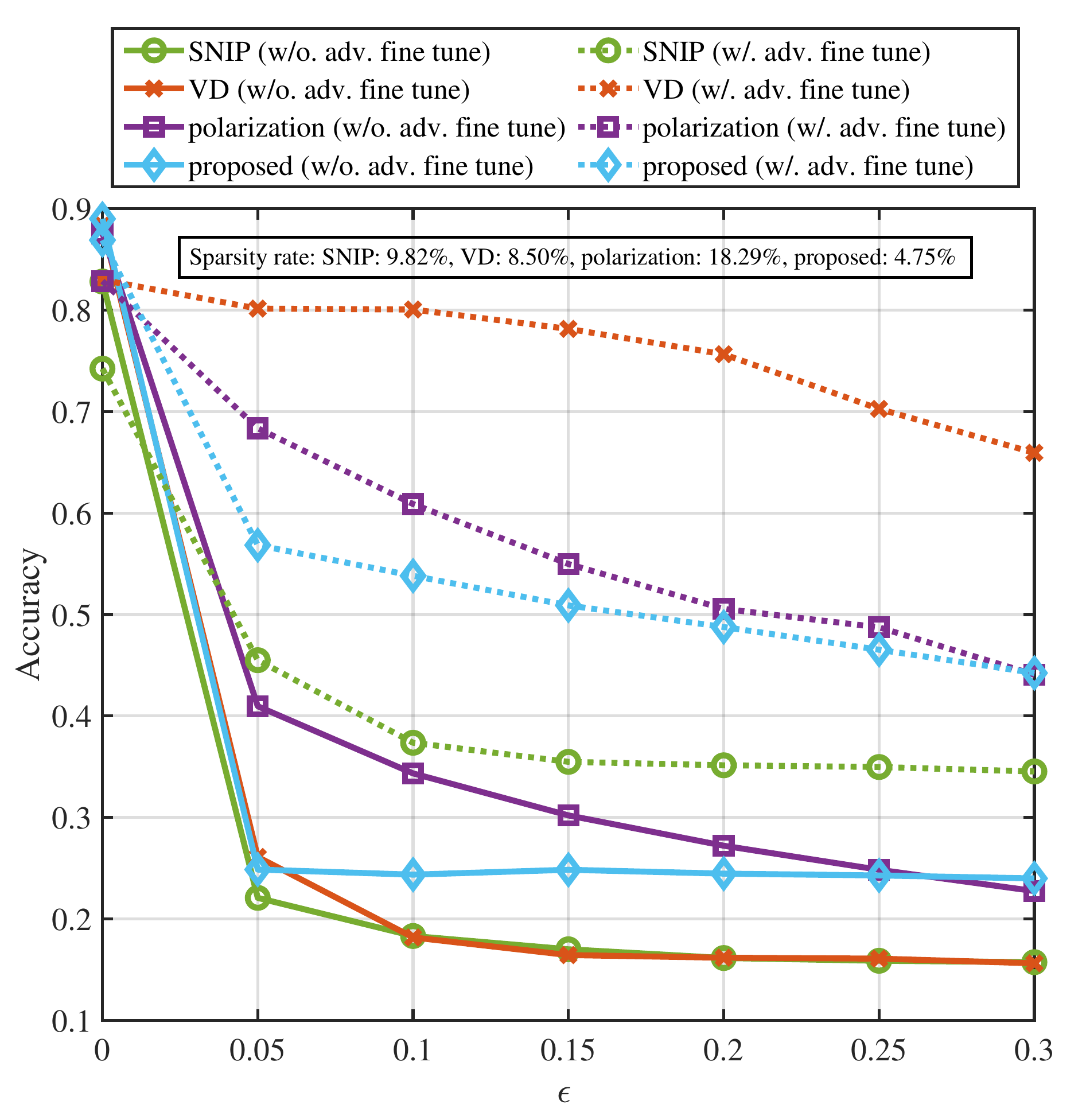}
\par\end{centering}
\caption{Accuracy vs $\epsilon$ for LeNet model.\textcolor{blue}{\label{fig:Accuracy-vs-epsilon}}}
\end{figure}

In this subsection, we evaluate the robustness of the pruned model.
The robustness of the pruned model is often neglected in model pruning
methods while it is of highly importance \cite{sehwag2020hydra,li2021robust,wang2018adversarial}.
We compare the model robustness of our proposed method to the baselines
on LeNet and Fashion-MNIST dataset. We generate 10000 adversarial
examples by fast gradient sign method (FGSM) and pass them through
the pruned models. In Fig. \ref{fig:Accuracy-vs-epsilon}, we plot
the model accuracy under different $\epsilon$. Both the accuracy
with and without an adversarial fine tuning is reported. It can be
observed that all methods show a rapid drop without fine tuning. Specifically,
our proposed method shows an accuracy drop of 72.1\% from 89.01\%
when $\epsilon=0$ to 24.86\% when $\epsilon=0.05$, while SNIP reports
the largest drop of 73.4\%. However, we also observe that the structured
pruning methods (proposed and polarization) show stronger robustness
than the unstructured ones (VD and SNIP). We think the reasons why
our proposed method and polarization perform better may be different.
For polarization, the unpruned weights are significantly more than
other baselines, which means the remaining important filters have
significantly more parameters. This allows the model to have a strong
ability to resist the noise. For our proposed method, the possible
reason is that it can enforce the weights to gather in groups. Thus,
the model can ``concentrate'' on some specific features even when
they are added by noise. It can be observed that after an adversarial
fine tuning, all methods show an improvement. VD shows the largest
improvement of 77.31\% at $\epsilon=0.1$ while our proposed method
only shows a moderate improvement of 56.26\%. The possible reason
behind this is our proposed method is ``too'' sparse and the remaining
weights cannot learn the new features. Note that SNIP shows the smallest
improvement. The possible reason is that in SNIP, the weights are
pruned before training and the auxiliary importance variables may
not measure the weight importance accurately, which results in some
important weights being pruned. Thus, the expressing ability of the
model may be harmed.

\section{Conclusions and Future Work\label{sec:Conclusions}}

\subsection*{Conclusions}

We considered the problem of model compression from a Bayesian perspective.
We first proposed a three-layer hierarchical sparse prior in which
an extra support layer is used to capture the desired structure for
the weights. Thus, the proposed sparse prior can promote both per-neuron
weight-level structured sparsity and neuron-level structured sparsity.
Then, we derived a Turbo-VBI based Bayesian compression algorithm
for the resulting model compression problem. Our proposed algorithm
has low complexity and high robustness. We further established the
convergence of the sparse VBI part in the Turbo-VBI algorithm. Simulation
results show that our proposed Turbo-VBI based model compression algorithm
can promote a more regular structure in the pruned neural networks
while achieving even better performance in terms of compression rate
and inferencing accuracy compared to the baselines.

\subsection*{Future Work}

With the proposed Turbo-VBI based structured model compression method,
we can promote more regular and more flexible structures and achieve
superior compressing performance during neural network pruning. This
also opens more possibilities for future research, as listed below:

Communication Efficient Federated Learning: Our proposed method shows
huge potential in a federated learning scenario. First, it can promote
a more regular structure in the weight matrix such that the weight
matrix has a lower entropy. Thus, the regular structure can be further
leveraged to reduce communication overhead in federated learning.
Second, the two-module design of the Turbo-VBI algorithm naturally
fits in a federated learning setting. The server can promote a global
sparse structure by running Module B while the clients can perform
local Bayesian training by running Module A.

Joint Design of Quantization and Pruning: Existing works \cite{Louizos2017,van2020bayesian}
have pointed out that quantization and pruning can be achieved simultaneously
by Bayesian model compression. Thus, it is also possible to further
compress the model by combining quantization with our proposed method.

\appendix{}

\subsection{Derivation of Three Subproblems}

Here, we provide the derivation from the original problem \eqref{eq:min_neg_ELBO}
to the three subproblems \eqref{eq:subproblem 1}, \eqref{eq:subproblem 2}
and \eqref{eq:subproblem 3}. For expression simplicity, let $\mathbf{z}=\left[\mathbf{z}_{1},\mathbf{z}_{2},\mathbf{z}_{3}\right]$
denote all the variables $\mathbf{w},\boldsymbol{\rho}$ and $\mathbf{s}$,
where $\mathbf{z}_{1}=\mathbf{w}$, $\mathbf{z}_{2}=\boldsymbol{\rho}$
and $\mathbf{z}_{3}=\mathbf{s}$. Then, the original problem \eqref{eq:min_neg_ELBO}
can be equivalently written as

\textbf{Sparse VBI Problem:}

\begin{equation}
\begin{aligned}\min_{q\left(\mathbf{z}\right)} & -ELBO,\\
s.t. & q\left(\mathbf{z}\right)=\prod_{n=1}^{3N}q\left(z_{n}\right).
\end{aligned}
\label{eq:min_neg_ELBO-1}
\end{equation}
 Then, by (15) in \cite{Tzikas2008}, we have 
\[
\begin{aligned}ELBO\\
= & \int q\left(\mathbf{z}\right)\ln\hat{p}\left(\mathcal{D}|\mathbf{z}\right)d\mathbf{z}-D_{KL}\left(q\left(\mathbf{z}\right)||\hat{p}\left(\mathbf{z}\right)\right)\\
= & \int q\left(\mathbf{z}\right)\ln\frac{\hat{p}\left(\mathcal{D},\mathbf{z}\right)}{q\left(\mathbf{z}\right)}d\mathbf{z}\\
= & \int\prod_{i=1}^{3}q\left(\mathbf{z}_{i}\right)\left[\ln\hat{p}\left(\mathcal{D},\mathbf{z}\right)-\sum_{i=1}^{3}\ln q\left(\mathbf{z}_{i}\right)\right]d\mathbf{z}\\
= & \int\prod_{i=1}^{3}q\left(\mathbf{z}_{i}\right)\ln\hat{p}\left(\mathcal{D},\mathbf{z}\right)\prod_{i=1}^{3}d\mathbf{z}_{i}\\
 & -\sum_{i=1}^{3}\int\prod_{j=1}^{3}q\left(\mathbf{z}_{j}\right)\ln q\left(\mathbf{z}_{i}\right)d\mathbf{z}_{i}\\
= & \int q\left(\mathbf{z}_{j}\right)\left[\ln\hat{p}\left(\mathcal{D},\mathbf{z}\right)\prod_{i\neq j}q\left(\mathbf{z}_{i}\right)d\mathbf{z}_{i}\right]d\mathbf{z}_{j}\\
 & -\int q\left(\mathbf{z}_{j}\right)\ln q\left(\mathbf{z}_{j}\right)d\mathbf{z}_{j}-\sum_{i\neq j}\int q\left(\mathbf{z}_{i}\right)\ln q\left(\mathbf{z}_{i}\right)d\mathbf{z}_{i}\\
= & \int q\left(\mathbf{z}_{j}\right)\ln\widetilde{p}_{j}d\mathbf{z}_{j}-\int q\left(\mathbf{z}_{j}\right)\ln q\left(\mathbf{z}_{j}\right)d\mathbf{z}_{j}\\
 & -\sum_{i\neq j}\int q\left(\mathbf{z}_{i}\right)\ln q\left(\mathbf{z}_{i}\right)d\mathbf{z}_{i}\\
= & -D_{KL}\left(q\left(\mathbf{z}_{j}\right)||\widetilde{p}_{j}\right)-\sum_{i\neq j}\int q\left(\mathbf{z}_{i}\right)\ln q\left(\mathbf{z}_{i}\right)d\mathbf{z}_{i},
\end{aligned}
\]
where $\ln\widetilde{p}_{j}=\int\ln\hat{p}\left(\mathcal{D},\mathbf{z}\right)\prod_{i\neq j}q\left(\mathbf{z}_{i}\right)d\mathbf{z}_{i}=\left\langle \hat{p}\left(\mathcal{D},\mathbf{z}\right)\right\rangle _{i\neq j}$.
It is obvious that $-ELBO$ is minimized when $D_{KL}\left(q\left(\mathbf{z}_{j}\right)||\widetilde{p}_{j}\right)$
is minimized. Thus, the Sparse VBI Problem \eqref{eq:min_neg_ELBO-1}
can be solved by iteratively minimizing $D_{KL}\left(q\left(\mathbf{z}_{j}\right)||\widetilde{p}_{j}\right)$
for $j=1,2,3$.

\subsection{Derivation of (25) - (30)}

Let $q^{\star}\left(\boldsymbol{\rho}\right)=\widetilde{p}_{\boldsymbol{\rho}}$,
we have 
\[
\begin{aligned}\ln q^{\star}\left(\boldsymbol{\rho}\right)\\
\propto & \ln\widetilde{p}_{\boldsymbol{\rho}}\\
\propto & \left\langle \ln p\left(\mathbf{w},\boldsymbol{\rho},\mathbf{s},\mathcal{D}\right)\right\rangle _{q\left(\mathbf{s}\right)q\left(\mathbf{w}\right)}\\
\propto & \left\langle \ln p\left(\mathbf{w}|\boldsymbol{\rho}\right)\right\rangle _{q\left(\mathbf{w}\right)}+\left\langle \ln p\left(\boldsymbol{\rho}|\mathbf{s}\right)\right\rangle _{q\left(\mathbf{s}\right)}\\
\propto & \sum_{n=1}^{N}\left(\left\langle s_{n}\right\rangle a_{n}+\left\langle 1-s_{n}\right\rangle \overline{a}_{n}\right)\ln\rho_{n}\\
 & -\left(\left\langle |w_{n}|^{2}\right\rangle +\left\langle s_{n}\right\rangle b_{n}+\left\langle 1-s_{n}\right\rangle \overline{b}_{n}\right)\rho_{n}.
\end{aligned}
\]
 Thus, we have that $q^{\star}\left(\boldsymbol{\rho}\right)$ follows
a Gamma distribution in (25), with parameters in (26) and (27).

Let $q^{\star}\left(\mathbf{s}\right)=\widetilde{p}_{\mathbf{s}}$,
we have
\[
\begin{aligned}\ln q^{\star}\left(\mathbf{s}\right)\\
\propto & \ln\widetilde{p}_{\mathbf{s}}\\
\propto & \left\langle \ln p\left(\mathbf{w},\boldsymbol{\rho},\mathbf{s},\mathcal{D}\right)\right\rangle _{q\left(\boldsymbol{\rho}\right)q\left(\mathbf{w}\right)}\\
\propto & \left\langle \ln p\left(\boldsymbol{\rho}|\mathbf{s}\right)\right\rangle _{q\left(\boldsymbol{\rho}\right)}+\ln\hat{p}\left(\mathbf{s}\right)\\
\propto & \sum_{n=1}^{N}s_{n}\left(\ln b_{n}^{a_{n}}+\left(a_{n}-1\right)\left\langle \ln\rho_{n}\right\rangle -b_{n}\left\langle \rho_{n}\right\rangle -\ln\Gamma\left(a_{n}\right)\right)
\end{aligned}
\]
\[
\begin{aligned}\qquad\quad\; & +\left(1-s_{n}\right)\left(\ln\overline{b}_{n}^{\overline{a}_{n}}+\left(\overline{a}_{n}-1\right)\left\langle \ln\rho_{n}\right\rangle -\overline{b}_{n}\left\langle \rho_{n}\right\rangle -\ln\Gamma\left(\overline{a}_{n}\right)\right)\\
 & +\sum_{n=1}^{N}\left(s_{n}\ln\pi_{n}+\left(1-s_{n}\right)\ln\left(1-\pi_{n}\right)\right)\\
\propto & \ln\prod_{n=1}^{N}\left(\widetilde{\pi}_{n}\right)^{s_{n}}\left(1-\widetilde{\pi}_{n}\right)^{1-s_{n}}.
\end{aligned}
\]

Thus, we have that $q^{\star}\left(\mathbf{s}\right)$ follows a Bernoulli
distribution in (28), with parameters in (29) and (30).


\begin{thebibliography}{10}
	\providecommand{\url}[1]{#1}
	\csname url@samestyle\endcsname
	\providecommand{\newblock}{\relax}
	\providecommand{\bibinfo}[2]{#2}
	\providecommand{\BIBentrySTDinterwordspacing}{\spaceskip=0pt\relax}
	\providecommand{\BIBentryALTinterwordstretchfactor}{4}
	\providecommand{\BIBentryALTinterwordspacing}{\spaceskip=\fontdimen2\font plus
		\BIBentryALTinterwordstretchfactor\fontdimen3\font minus
		\fontdimen4\font\relax}
	\providecommand{\BIBforeignlanguage}[2]{{%
			\expandafter\ifx\csname l@#1\endcsname\relax
			\typeout{** WARNING: IEEEtran.bst: No hyphenation pattern has been}%
			\typeout{** loaded for the language `#1'. Using the pattern for}%
			\typeout{** the default language instead.}%
			\else
			\language=\csname l@#1\endcsname
			\fi
			#2}}
	\providecommand{\BIBdecl}{\relax}
	\BIBdecl
	
	\bibitem{Lee2018}
	N.~Lee, T.~Ajanthan, and P.~H. Torr, ``Snip: Single-shot network pruning based
	on connection sensitivity,'' \emph{International Conference on Learning
		Representations (ICLR)}, 2019.
	
	\bibitem{LeCun1990}
	Y.~LeCun, J.~S. Denker, and S.~A. Solla, ``Optimal brain damage,'' in
	\emph{Advances in Neural Information Processing Systems (NeurIPS)}, 1990, pp.
	598--605.
	
	\bibitem{Hassibi1993}
	B.~Hassibi and D.~G. Stork, \emph{Second Order Derivatives for Network Pruning:
		Optimal Brain Surgeon}.\hskip 1em plus 0.5em minus 0.4em\relax Morgan
	Kaufmann, 1993.
	
	\bibitem{Louizos2017a}
	C.~Louizos, M.~Welling, and D.~P. Kingma, ``Learning sparse neural networks
	through $l\_0$ regularization,'' \emph{arXiv preprint arXiv:1712.01312},
	2017.
	
	\bibitem{Wen2016}
	W.~Wen, C.~Wu, Y.~Wang, Y.~Chen, and H.~Li, ``Learning structured sparsity in
	deep neural networks,'' \emph{arXiv preprint arXiv:1608.03665}, 2016.
	
	\bibitem{Neill2020}
	J.~O. Neill, ``An overview of neural network compression,'' \emph{arXiv
		preprint arXiv:2006.03669}, 2020.
	
	\bibitem{Scardapane2017}
	S.~Scardapane, D.~Comminiello, A.~Hussain, and A.~Uncini, ``Group sparse
	regularization for deep neural networks,'' \emph{Neurocomputing}, vol. 241,
	pp. 81--89, 2017.
	
	\bibitem{howard2019searching}
	A.~Howard, M.~Sandler, G.~Chu, L.-C. Chen, B.~Chen, M.~Tan, W.~Wang, Y.~Zhu,
	R.~Pang, V.~Vasudevan \emph{et~al.}, ``Searching for mobilenetv3,'' in
	\emph{Proceedings of the IEEE/CVF International Conference on Computer Vision
		(ICCV)}, 2019, pp. 1314--1324.
	
	\bibitem{zhuang2020neuron}
	T.~Zhuang, Z.~Zhang, Y.~Huang, X.~Zeng, K.~Shuang, and X.~Li, ``Neuron-level
	structured pruning using polarization regularizer,'' \emph{Advances in Neural
		Information Processing Systems (NeurIPS)}, vol.~33, pp. 9865--9877, 2020.
	
	\bibitem{yang2020harmonious}
	L.~Yang, Z.~He, and D.~Fan, ``Harmonious coexistence of structured weight
	pruning and ternarization for deep neural networks,'' in \emph{Proceedings of
		the AAAI Conference on Artificial Intelligence}, vol.~34, no.~04, 2020, pp.
	6623--6630.
	
	\bibitem{Weigend1991}
	A.~S. Weigend, D.~E. Rumelhart, and B.~A. Huberman, ``Generalization by
	weight-elimination with application to forecasting,'' in \emph{Advances in
		Neural Information Processing Systems (NeurIPS)}, 1991, pp. 875--882.
	
	\bibitem{frankle2018lottery}
	J.~Frankle and M.~Carbin, ``The lottery ticket hypothesis: Finding sparse,
	trainable neural networks,'' \emph{arXiv preprint arXiv:1803.03635}, 2018.
	
	\bibitem{frankle2020linear}
	J.~Frankle, G.~K. Dziugaite, D.~Roy, and M.~Carbin, ``Linear mode connectivity
	and the lottery ticket hypothesis,'' in \emph{International Conference on
		Machine Learning (ICML)}.\hskip 1em plus 0.5em minus 0.4em\relax PMLR, 2020,
	pp. 3259--3269.
	
	\bibitem{van2020bayesian}
	M.~Van~Baalen, C.~Louizos, M.~Nagel, R.~A. Amjad, Y.~Wang, T.~Blankevoort, and
	M.~Welling, ``Bayesian bits: Unifying quantization and pruning,''
	\emph{Advances in Neural Information Processing Systems (NeurIPS)}, vol.~33,
	pp. 5741--5752, 2020.
	
	\bibitem{Molchanov2017}
	D.~Molchanov, A.~Ashukha, and D.~Vetrov, ``Variational dropout sparsifies deep
	neural networks,'' in \emph{International Conference on Machine Learning
		(ICML)}.\hskip 1em plus 0.5em minus 0.4em\relax PMLR, 2017, pp. 2498--2507.
	
	\bibitem{Louizos2017}
	C.~Louizos, K.~Ullrich, and M.~Welling, ``Bayesian compression for deep
	learning,'' \emph{arXiv preprint arXiv:1705.08665}, 2017.
	
	\bibitem{Liu2020}
	A.~Liu, G.~Liu, L.~Lian, V.~K. Lau, and M.-J. Zhao, ``Robust recovery of
	structured sparse signals with uncertain sensing matrix: A turbo-vbi
	approach,'' \emph{IEEE Trans. Wireless Commun.}, vol.~19, no.~5, pp.
	3185--3198, 2020.
	
	\bibitem{Schniter2010}
	P.~Schniter, ``Turbo reconstruction of structured sparse signals,'' in
	\emph{2010 44th Annual Conference on Information Sciences and Systems
		(CISS)}.\hskip 1em plus 0.5em minus 0.4em\relax IEEE, 2010, pp. 1--6.
	
	\bibitem{Wu2018}
	A.~Wu, S.~Nowozin, E.~Meeds, R.~E. Turner, J.~M. Hernandez-Lobato, and A.~L.
	Gaunt, ``Deterministic variational inference for robust {Bayesian} neural
	networks,'' \emph{arXiv preprint arXiv:1810.03958}, 2018.
	
	\bibitem{zhang2019eager}
	J.~Zhang, X.~Chen, M.~Song, and T.~Li, ``Eager pruning: Algorithm and
	architecture support for fast training of deep neural networks,'' in
	\emph{2019 ACM/IEEE 46th Annual International Symposium on Computer
		Architecture (ISCA)}.\hskip 1em plus 0.5em minus 0.4em\relax IEEE, 2019, pp.
	292--303.
	
	\bibitem{Friedman2010}
	J.~Friedman, T.~Hastie, and R.~Tibshirani, ``A note on the group lasso and a
	sparse group lasso,'' \emph{arXiv preprint arXiv:1001.0736}, 2010.
	
	\bibitem{yang2017designing}
	T.-J. Yang, Y.-H. Chen, and V.~Sze, ``Designing energy-efficient convolutional
	neural networks using energy-aware pruning,'' in \emph{Proceedings of the
		IEEE Conference on Computer Vision and Pattern Recognition (CVPR)}, 2017, pp.
	5687--5695.
	
	\bibitem{kwon2020structured}
	S.~J. Kwon, D.~Lee, B.~Kim, P.~Kapoor, B.~Park, and G.-Y. Wei, ``Structured
	compression by weight encryption for unstructured pruning and quantization,''
	in \emph{Proceedings of the IEEE/CVF Conference on Computer Vision and
		Pattern Recognition (CVPR)}, 2020, pp. 1909--1918.
	
	\bibitem{Yu2017}
	J.~Yu, A.~Lukefahr, D.~Palframan, G.~Dasika, R.~Das, and S.~Mahlke, ``Scalpel:
	Customizing dnn pruning to the underlying hardware parallelism,'' in
	\emph{2017 ACM/IEEE 44th Annual International Symposium on Computer
		Architecture (ISCA)}, 2017, pp. 548--560.
	
	\bibitem{blalock2020state}
	D.~Blalock, J.~J. Gonzalez~Ortiz, J.~Frankle, and J.~Guttag, ``What is the
	state of neural network pruning?'' \emph{Proceedings of Machine Learning and
		Systems}, vol.~2, pp. 129--146, 2020.
	
	\bibitem{Jospin2020}
	L.~V. Jospin, W.~Buntine, F.~Boussaid, H.~Laga, and M.~Bennamoun, ``Hands-on
	{Bayesian} neural networks--a tutorial for deep learning users,'' \emph{arXiv
		preprint arXiv:2007.06823}, 2020.
	
	\bibitem{Murphy2013}
	K.~Murphy, Y.~Weiss, and M.~I. Jordan, ``Loopy belief propagation for
	approximate inference: An empirical study,'' \emph{arXiv preprint
		arXiv:1301.6725}, 2013.
	
	\bibitem{Tzikas2008}
	D.~G. Tzikas, A.~C. Likas, and N.~P. Galatsanos, ``The variational
	approximation for {Bayesian} inference,'' \emph{IEEE Signal Processing Mag.},
	vol.~25, no.~6, pp. 131--146, 2008.
	
	\bibitem{Kingma2015a}
	D.~P. Kingma, T.~Salimans, and M.~Welling, ``Variational dropout and the local
	reparameterization trick,'' \emph{Advances in Neural Information Processing
		Systems (NeurIPS)}, vol.~28, pp. 2575--2583, 2015.
	
	\bibitem{Tseng2001}
	P.~Tseng, ``Convergence of a block coordinate descent method for
	nondifferentiable minimization,'' \emph{J. Optim. Theory Appl.}, vol. 109,
	no.~3, pp. 475--494, 2001.
	
	\bibitem{yuan2019enhanced}
	X.~Yuan, L.~Ren, J.~Lu, and J.~Zhou, ``Enhanced bayesian compression via deep
	reinforcement learning,'' in \emph{Proceedings of the IEEE/CVF Conference on
		Computer Vision and Pattern Recognition (CVPR)}, 2019, pp. 6946--6955.
	
	\bibitem{xiao2017fashion}
	H.~Xiao, K.~Rasul, and R.~Vollgraf, ``Fashion-mnist: a novel image dataset for
	benchmarking machine learning algorithms,'' \emph{arXiv preprint
		arXiv:1708.07747}, 2017.
	
	\bibitem{KrizhevskyaccessedAugust212022}
	A.~Krizhevsky, V.~Nair, and G.~Hinton, ``The cifar-10 and cifar-100 dataset.''
	\emph{cs.toronto.edu. http://www.cs.toronto.edu/$~$kriz/cifar.html},
	(accessed August 21 2022).
	
	\bibitem{sehwag2020hydra}
	V.~Sehwag, S.~Wang, P.~Mittal, and S.~Jana, ``Hydra: Pruning adversarially
	robust neural networks,'' \emph{Advances in Neural Information Processing
		Systems (NeurIPS)}, vol.~33, pp. 19\,655--19\,666, 2020.
	
	\bibitem{li2021robust}
	J.~Li, R.~Drummond, and S.~R. Duncan, ``Robust error bounds for quantised and
	pruned neural networks,'' in \emph{Learning for Dynamics and Control}.\hskip
	1em plus 0.5em minus 0.4em\relax PMLR, 2021, pp. 361--372.
	
	\bibitem{wang2018adversarial}
	L.~Wang, G.~W. Ding, R.~Huang, Y.~Cao, and Y.~C. Lui, ``Adversarial robustness
	of pruned neural networks,'' 2018.
	
\end{thebibliography}


\end{document}